\documentclass[10pt,twocolumn,letterpaper]{article}

\usepackage[pagenumbers]{cvpr} % To force page numbers, e.g. for an arXiv version

\usepackage{graphicx}
\usepackage{capt-of}
\usepackage{amsmath,amssymb,amsfonts,dsfont,bm,bbm,mathrsfs,pifont}
\usepackage{booktabs,multirow,adjustbox}
\usepackage[pagebackref,breaklinks,colorlinks]{hyperref}
\usepackage{cleveref}  % Should be loaded after 'hyperref', and works perfectly with 'subfigure'.
\crefname{section}{Sec.}{Secs.}
\Crefname{section}{Section}{Sections}
\crefname{table}{Tab.}{Tabs.}
\Crefname{table}{Table}{Tables}
\crefname{figure}{Fig.}{Figs.}
\Crefname{figure}{Figure}{Figures}
\crefname{equation}{Eq.}{Eqs.}
\Crefname{equation}{Equation}{Equations}
\newcommand{\FID}{FID$\downarrow$}           % Notation of FID score in Table.
\newcommand{\FIDD}{FID (Depth)$\downarrow$}  % Notation of FID(Depth) score in Table.
\newcommand{\RPE}{RP$\downarrow$}           % Notation of RPE score in Table.
\newcommand{\RCE}{RC$\downarrow$}           % Notation of RCE score in Table.
\newcommand{\RealD}{DP (Real)$\downarrow$}      % Notation of real depth
\newcommand{\FakeD}{DP (Fake)$\downarrow$}      % Notation of fake depth
\newcommand{\tb}[1]{\textbf{#1}}

\def\cvprPaperID{5203}
\def\confName{CVPR}
\def\confYear{2022}

\begin{document}

%%%% Title
% \title{Depth-Guided GAN for 3D-Aware Indoor Scene Synthesis}
\title{3D-Aware Indoor Scene Synthesis with Depth Priors}

%%%% Authors
\author{
    Zifan Shi$^\dagger$ \quad
    Yujun Shen$^\ddagger$ \quad
    Jiapeng Zhu$^\dagger$ \quad
    Dit-Yan Yeung$^\dagger$ \quad
    Qifeng Chen$^\dagger$ \\
    $^\dagger$The Hong Kong University of Science and Technology \quad
    $^\ddagger$ByteDance Inc.
}

\maketitle

\begin{abstract}
Despite the recent advancement of Generative Adversarial Networks (GANs) in learning 3D-aware image synthesis from 2D data, existing methods fail to model indoor scenes due to the large diversity of room layouts and the objects inside.
%
% under such a complex scenario
%
We argue that indoor scenes do not have a shared intrinsic structure, and hence only using 2D images cannot adequately guide the model with the 3D geometry.
In this work, we fill in this gap by introducing depth as a 3D prior.
Compared with other 3D data formats, depth better fits the convolution-based generation mechanism and is more easily accessible in practice.
%
% considering its easy access compared with other data formats, like point cloud and implicit surface.
%
Specifically, we propose a dual-path generator, where one path is responsible for depth generation, whose intermediate features are injected into the other path as the condition for appearance rendering.
Such a design eases the 3D-aware synthesis with explicit geometry information.
Meanwhile, we introduce a switchable discriminator both to differentiate real \textit{v.s.} fake domains and to predict the depth from a given input.
In this way, the discriminator can take the spatial arrangement into account and advise the generator to learn an appropriate depth condition.
%
% With such a design, we can adequately capture the depth-appearance relationship, making the model better aware of the 3D geometry.
%
Extensive experimental results suggest that our approach is capable of synthesizing indoor scenes with impressively good quality and 3D consistency, significantly outperforming state-of-the-art alternatives.%
\footnote{Project page can be found \href{https://vivianszf.github.io/depthgan}{here}.}
\end{abstract}

% Despite the recent advancement of Generative Adversarial Networks (GANs) in learning 3D-aware image synthesis from 2D data, existing methods fail to model indoor scenes due to the large diversity of room layouts and the objects inside. We argue that indoor scenes do not have a shared intrinsic structure, and hence only using 2D images cannot adequately guide the model with the 3D geometry. In this work, we fill in this gap by introducing depth as a 3D prior. Compared with other 3D data formats, depth better fits the convolution-based generation mechanism and is more easily accessible in practice. Specifically, we propose a dual-path generator, where one path is responsible for depth generation, whose intermediate features are injected into the other path as the condition for appearance rendering. Such a design eases the 3D-aware synthesis with explicit geometry information. Meanwhile, we introduce a switchable discriminator to both differentiate real v.s. fake domains and predict the depth from a given input. In this way, the discriminator can take the spatial arrangement into account and advise the generator to learn an appropriate depth condition. Extensive experimental results suggest that our approach is capable of synthesizing indoor scenes with impressively good quality and 3D consistency, significantly outperforming state-of-the-art alternatives.
\section{Introduction}\label{sec:intro}

\begin{figure}[t]
    \centering
    \includegraphics[width=1.0\linewidth]{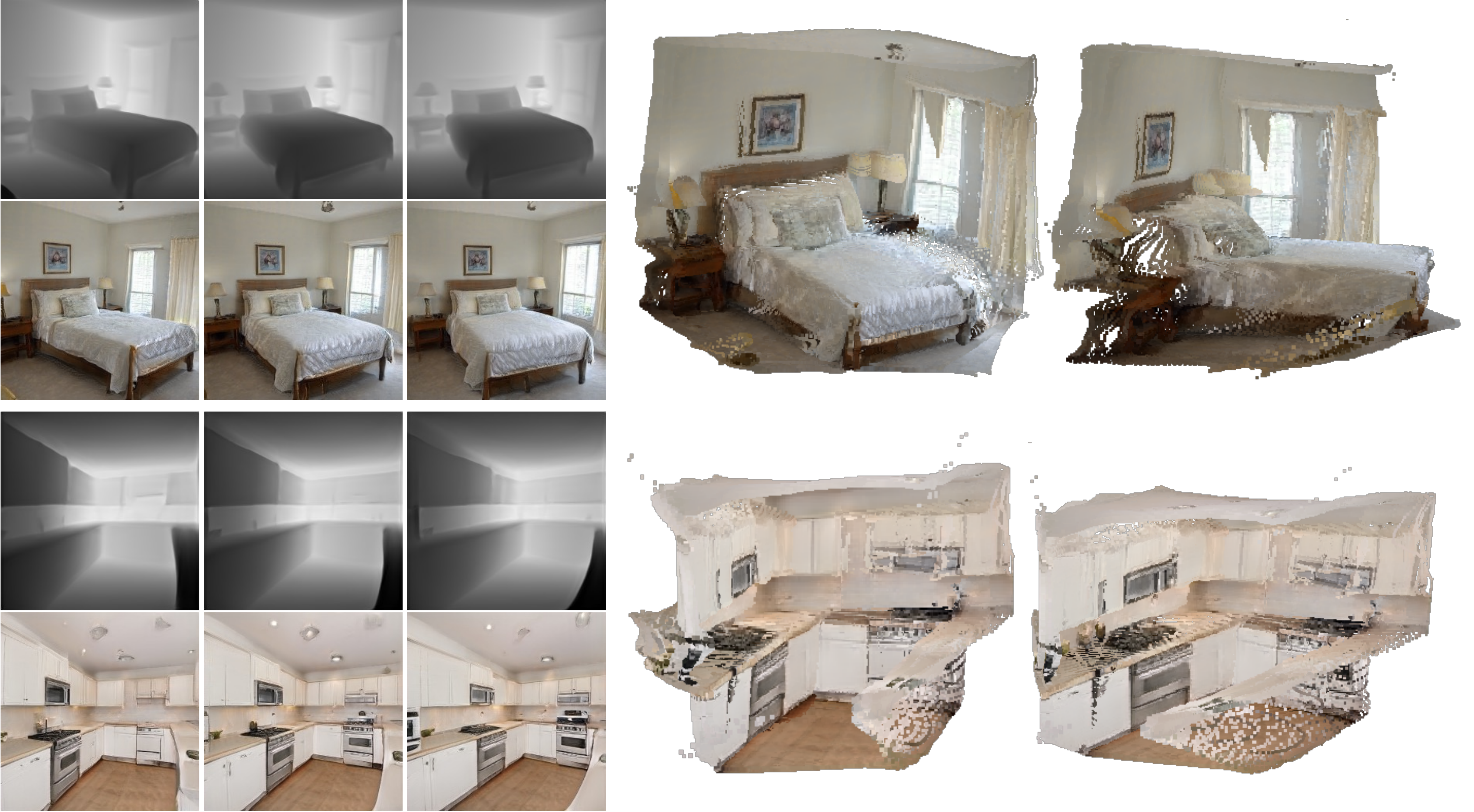}
    \vspace{-15pt}
    \caption{
        \textbf{Photo-realistic bedroom and kitchen produced by our DepthGAN.}
        Left: Two sets of synthesized depth maps and their corresponding rendered images from three different viewpoints.
        Right: Visualization using ~\cite{Zhou2018} of the 3D reconstruction results from above synthesized samples.
        % From left to right: three synthesized depth maps and their corresponding rendered images from different viewpoints, 3D visualization on the synthesized results from two different viewpoints using~\cite{Zhou2018}.
    }
    \label{fig:teaser}
    \vspace{-10pt}
\end{figure}

% \begin{figure}[t]
%     \centering
%     \includegraphics[width=1.0\linewidth]{Figs/teaser.pdf}
%     \vspace{-20pt}
%     \caption{
%         \textbf{Photo-realistic bedroom and kitchen produced by our DepthGAN.}
%         %
%         From left to right: the synthesized depths, the corresponding rendered images, 3D visualization on the synthesized results from two different viewpoints using~\cite{Zhou2018}.
%     }
%     \label{fig:teaser}
%     \vspace{-10pt}
% \end{figure}

Generative Adversarial Networks (GANs)~\cite{gan} have enabled high-fidelity 2D image synthesis, but how to make a GAN model aware of 3D information remains unsolved.
Along with the recent advent of Neural Radiance Field (NeRF)~\cite{nerf} for 3D scene reconstruction, some attempts~\cite{piGAN2021, graf, GIRAFFE} propose to incorporate NeRF into GANs to learn a 3D-aware image generator from a 2D image collection.
Instead of using 2D convolutional layers, the generator is asked to learn a point-wise implicit function, which maps the 3D coordinates to volume densities and colors~\cite{nerf, graf}.
% Concretely, the generator produces images with volume rendering instead of 2D convolutional layers, while the discriminator differentiates real and fake samples \textit{still} from the 2D image space, same as conventional 2D GANs.

Although existing methods show promising results in learning 3D-aware object synthesis, such as human faces and cars, they exhibit severe performance degradation on indoor scene datasets, such as bedrooms and kitchens.
There are mainly two reasons.
First, objects normally have a shared intrinsic structure, which eases the difficulty of modeling 3D geometry from 2D images \textit{only}.
For instance, human heads share similar shapes, and each face consists of two eyes located at relatively defined positions.
On the contrary, indoor scenes have much higher diversity, considering the complex room layout and the interior decoration~\cite{higan}.
Second, existing methods assume the distribution of camera poses~\cite{graf, campari}.
Such an assumption is sound under the case of object synthesis because objects are commonly placed at the center of a 2D image.
Indoor scene images are usually shot from far more diverse viewpoints, making it too challenging for the NeRF-based approaches to handle.

In this work, we propose a new paradigm for 3D-aware image synthesis by explicitly introducing a 3D prior into 2D GANs.
Compared with the volume renderer equipped with Multi-Layer Perceptron (MLP)~\cite{graf, piGAN2021, GIRAFFE}, GANs built on Convolutional Neural Network (CNN) achieve much more appealing synthesis performances~\cite{stylegan, stylegan2, stylegan3}, especially from the image quality and the image resolution perspectives.
Among numerous 3D data formats, such as point cloud~\cite{schnabel2007efficient,rusu20113d}, voxel~\cite{ashburner2000voxel, cheung2000real}, and implicit surface~\cite{ohtake2004ridge, michalkiewicz2019implicit}, we choose depth as our prior as it is defined in the 2D domain and hence naturally suitable for the convolution-based generator.
In addition, there are many publicly available depth datasets~\cite{nyu, cityscapes, kitti} and depth predictors~\cite{leres2021, midas}, making depth data easily accessible in practice.
%
% due to its easiest access in practice (\textit{e.g.}, depth can be either collected with the RGBD camera or predicted from RGB images through a deep model).

To sufficiently leverage the depth prior, we re-design the objectives of both the generator and the discriminator in a conventional GAN.
For one thing, we ask the generator to synthesize a 2D image accompanied by its corresponding depth.
To meet this goal, we carefully tailor a dual-path architecture, where the appearance-path takes the multi-level feature maps from the depth-path as the input conditions.
Through such a design, we manage to explicitly inject the geometry information into the generator.
For another, unlike the conventional discriminator that makes the real/fake decision from the 2D space, we learn a 3D-aware switchable discriminator.
Specifically, it is asked to distinguish the real and synthesized samples based on the image-depth joint distribution and, simultaneously, predict the depth from an input image.
The depth prediction is trained on real data and further used to supervise the fake data.
In this way, the discriminator is able to gain more knowledge on the spatial layout and better guide the generator from the 3D perspective.

We evaluate our approach, termed as \textbf{DepthGAN}, on a couple of challenging indoor scene datasets.
Both qualitative and quantitative results demonstrate the sufficient superiority of DepthGAN over existing methods.
For example, we improve Fréchet Inception Distance (FID)~\cite{fid} from $44.232$ to $4.797$ on LSUN bedroom dataset~\cite{yu2015lsun} in $256\times256$ resolution. 3D visualization on a set of synthesized images is shown in \cref{fig:teaser}.

% To summarize, our main \textit{contributions} are as follows.
% %
% (1) We propose a new paradigm for 3D-aware image synthesis by integrating depth as a prior in addition to 2D images.
% %
% (2) We develop \textbf{DepthGAN}, consisting of a dual-path generator and a switchable discriminator, to adequately utilize the depth prior for image generation.
% %
% (3) Our approach enables 3D-aware indoor scene synthesis with high image quality and good 3D controllability, significantly surpassing existing competitors both qualitatively and quantitatively.

\section{Related Work}\label{sec:related}

\noindent\textbf{GAN-based Image Synthesis.}
With the advent of Generative Adversarial Networks (GANs)~\cite{gan}, a large number of works have been proposed to generate high-quality photorealistic images~\cite{pggan, stylegan, stylegan2, biggan}.
To gain explicit control of the images, researchers study the disentanglement of different properties such as poses.
Supervised methods~\cite{shen2020interfacegan} leverage off-the-shelf attribute classifiers or image transformations to annotate the synthesized data and use the labeled data to guide the subspace learning in the latent space.
Unsupervised methods~\cite{ganspace, shen2021closed,peng2021} learn the control by analyzing the statistics or the model weights.
While these works can control the poses with the azimuth and elevation angles, the changes may violate the intrinsic consistency in the 3D space since there is no such constraint.

\noindent\textbf{3D-Aware Image Synthesis.}
Realizing that previous image synthesis methods do not consider 3D geometry, a large number of works have started to add 3D constraints for image synthesis.
Voxel-based methods like HoloGAN~\cite{HoloGAN2019} and BlockGAN~\cite{BlockGAN2020} learn low-dimensional 3D representations with deep voxels, followed by a learnable 3D-to-2D projection.
Inspired by NeRF~\cite{nerf}, GRAF~\cite{graf} and pi-GAN~\cite{piGAN2021} utilize the expensive rendering procedure to enforce the generative models to learn the 3D consistency and yield higher-quality 2D images.
GIRAFFE~\cite{GIRAFFE} proposes compositional neural radiance for image rendering and accelerates the rendering process with ConvNets.
StyleNeRF~\cite{gu2021stylenerf} proposes the NeRF path regularization to enforce the output to be closer to the output of NeRF that has multi-view consistency.
RGBD-GAN~\cite{RGBDGAN} samples two camera parameters to synthesize RGBD images from two views and then warps them to each other to ensure 3D consistency. Unlike RGBD-GAN, we synthesize depth images conditioned on RGB images, which are treated with unequal status.
All the works mentioned above learn geometry and appearance from 2D RGB images alone.
Due to the complexity of 3D geometry modeling and the lack of explicit 3D information, they target objects or well-aligned scenes and fail to generate high-quality images for complex scenes like bedrooms and kitchens.
In contrast, some other works utilize 3D prior knowledge to facilitate the learning of 3D consistency.
3D-GAN~\cite{3dgan}, VON~\cite{VON} and NGP~\cite{chen2021ngp} select shape as the 3D prior and use the expensive 3D-conv-based GAN to learn the geometry information that is costly and unable to model fine details of the shape.
%
% In addition to the shape prior, NGP uses albedo maps and normal maps as well, resulting in multiple 2D GANs to learn all the 3D attributes.
%
% GIS~\cite{2018GIS} assumes that normal map, material segmentation map and depths map are given and the appearance is generated accordingly.
% 
NGP~\cite{chen2021ngp} and GIS~\cite{2018GIS} utilize more than one 3D prior, such as albedo maps and normal maps, resulting in multiple 2D GANs to learn all the 3D attributes.
Instead of generating objects only, S$^2$-GAN~\cite{SSGAN2016} synthesizes indoor scenes with the help of normal maps, but it adopts the two-stage training to learn geometry first and the appearance next.
%
% Apart from the common 2D GAN for image synthesis, they employ an extra FCN to predict the surface normal maps to enforce the generator to learn the 3D information.
%
All the mentioned works either have separate 3D and 2D discriminators which learn the geometry and texture distributions independently, or use 2D discriminators only to make the real/fake decision on one 3D attribute or the appearance.
%
% Therefore, 3D knowledge and 2D knowledge are split for the discriminator.
%
In contrast, our discriminator is endowed with 3D and 2D knowledge at the same time.
GSN~\cite{GSN} follows the NeRF rendering structure and adds another depth channel in the discriminator to incorporate 3D priors, but it fails to generate images with a large diversity and reasonable fidelity due to the complex rendering process, the special requirements for training data, and inadequacy of its discriminator.

\section{Method}\label{sec:method}

\begin{figure*}[t]
    \centering
    \includegraphics[width=1.0\linewidth]{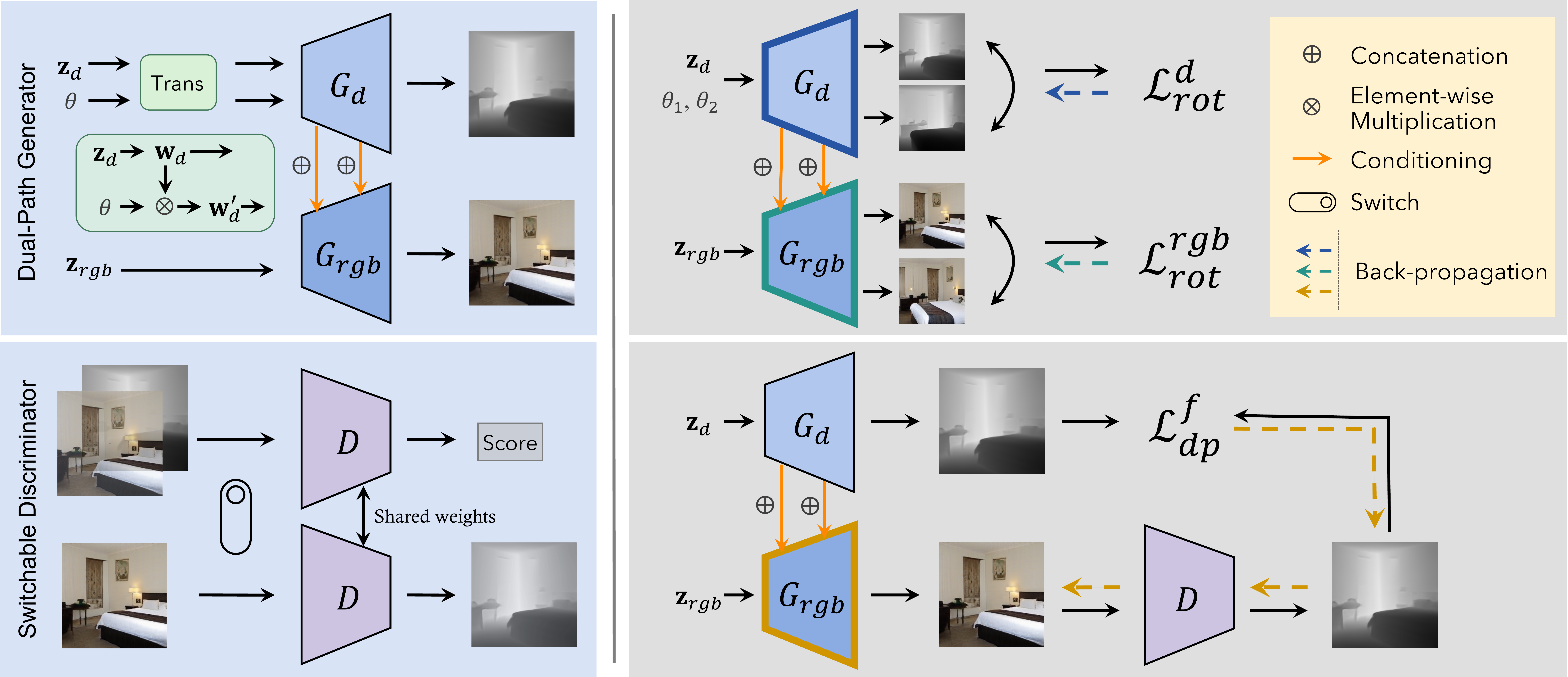}
    \vspace{-20pt}
    \caption{
        \textbf{The framework of DepthGAN}, consisting of a \textit{dual-path} generator that takes in two latent codes to generate the RGBD image with the appearance conditioned on the geometry, and a \textit{switchable} discriminator that produces the realness score from an RGBD image and predicts the depth map from an RGB image.
        Black arrows indicate the forward computation, while dashed arrows under different colors stand for the back-propagation regarding different objective functions. 
    }
    \label{fig:framework}
    \vspace{-15pt}
\end{figure*}

In this work, we propose a new diagram for 3D-aware image synthesis via introducing depth as a 3D prior into 2D GANs.
To adequately use the depth prior, we re-design both the generator and the discriminator in conventional GANs~\cite{gan}.
Concretely, we propose a \textit{dual-path} generator and a \textit{switchable} discriminator based on the recent StyleGAN2~\cite{stylegan2} model.
The overall framework is shown in \cref{fig:framework}.
For simplicity, we denote the RGB image, RGBD image, and depth image with $\mathbf{I}_{rgb}$, $\mathbf{I}_{rgbd}$, and $\mathbf{I}_{d}$, respectively.

\subsection{Dual-Path Generator}

To make the generator become aware of the geometry information, we ask it to synthesize a depth image accompanied with the RGB image.
For this purpose, we tailor a dual-path generator, consisting of a depth generator $G_d$ and an appearance renderer $G_{rgb}$.
Two latent spaces, $\mathcal{Z}_d$ and $\mathcal{Z}_{rgb}$, are introduced to enable the independent sampling of depth and appearance.
To make sure the appearance is properly rendered on top of the geometry, we feed the intermediate feature maps of $G_d$ into $G_{rgb}$ as the condition.

\vspace{5pt}
\noindent\textbf{Depth Generator.}
To control the viewing point of the generated depth, we uniformly sample an angle $\theta$ from $[\theta_L, \theta_R]$.
Since networks tend to learn better information from high-frequency signals~\cite{sitzmann2019siren}, we encode $\theta$ with
\begin{align}
    \gamma(\theta, t) = h(sin(\theta), cos(\theta),..., sin(t\theta), cos(t\theta)),  \label{eq:angle-encoding}
\end{align}
where $t$ determines the maximum frequency.
$h:\mathbb{R}^{2t}\rightarrow \mathbb{R}^{m}$ stands for a non-linear mapping, which is implemented by a two-layer fully-connection (FC).
Like StyleGAN2~\cite{stylegan2}, the raw depth latent code $\mathbf{z}_d \in \mathcal{Z}_d$ is projected into a more disentangled latent space, resulting in $\mathbf{w}_d \in \mathbb{R}^{m}$.
The angle information is then injected to $\mathbf{w}_d$ through
\begin{align}
    \mathbf{w}'_d = \mathbf{w}_d \otimes \gamma(\theta, t),  \label{eq:angle-injection}
\end{align}
where $\otimes$ denotes the element-wise multiplication.
$\mathbf{w}'_d$ guides $G_d$ on synthesizing the depth image, $\mathbf{I}^f_d$, via layer-wise style modulation~\cite{stylegan2}.
Note that only the first two layers of $G_d$ employ $\mathbf{w}'_d$ while the remaining layers still use $\mathbf{w}_d$, because only early layers correspond to the viewing point of the output image~\cite{higan}.

%It introduces an inductive bias to learn 3D shape representations in canonical orientations which otherwise would be arbitrary. From GIRAFFE, to be checked

\vspace{5pt}
\noindent\textbf{Depth-Conditioned Appearance Renderer.}
$G_{rgb}$ shares a similar structure as $G_{d}$ with three modifications.
First, the number of output channels is 3 ($\mathbf{I}^f_{rgb}$) instead of 1 ($\mathbf{I}^f_d$).
Second, $G_{rgb}$ does not take the angle $\theta$ as the input.
Third, most importantly, $G_{rgb}$ takes the intermediate feature maps of $G_d$ as the conditions to acquire the geometry information.
Specifically, we first concatenate the per-layer feature $\mathbf{\Psi}_i$ of $G_d$ with that of $G_{rgb}$, $\mathbf{\Phi}_i$.
Here, $i$ denotes the layer index.
We then transform the concatenated result with
\begin{align}
    \mathbf{\Phi}'_i = f(\mathbf{\Psi}_i \oplus \mathbf{\Phi}_i),  \label{eq:depth-condition}
\end{align}
where $\oplus$ stands for the concatenation operation, and $f$ is implemented with a two-layer convolution.
$\mathbf{\Phi}'_i$ has the same number of channels as $\mathbf{\Phi}_i$.

\subsection{Switchable Discriminator}

Unlike the discriminator in conventional GANs that simply differentiates the real and fake domains from the RGB image space, we propose a switchable discriminator to compete with the generator by taking the spatial arrangement into account.
This is achieved from two aspects.
On one hand, $D$ makes the real/fake decision based on the joint distribution of RGB images and the corresponding depths.
In other words, $D$ takes an RGBD image as the input and outputs the realness score.
On the other hand, to better capture the relationship between the image and the depth, we ask $D$ to predict the depth from a given RGB image.
Concretely, we introduce a separate branch on top of some intermediate feature maps of $D$ for depth prediction.
Detailed structure of the depth prediction branch can be found in \textit{Appendix}.

To summarize, $D$ switches between the 4-channel RGBD inputs (\textit{i.e.}, for realness discrimination) and the 3-channel RGB inputs (\textit{i.e.}, for depth prediction).
To achieve this goal, we come up with a switchable input layer that adaptively adjusts the number of convolutional kernels.

\subsection{Training Objectives}

\noindent\textbf{Adversarial Loss.}
We adopt the standard adversarial loss for GAN training
\begin{align}
    \mathcal{L}^d_{adv} & = - \mathbb{E}[\log(D(\mathbf{I}^r_{rgbd}))] - \mathbb{E}[\log(1 - D(\mathbf{I}^f_{rgbd}))], \label{eq:loss_d} \\
    \mathcal{L}^g_{adv} & = - \mathbb{E}[\log(D(\mathbf{I}^f_{rgbd}))], \label{eq:loss_g}
    % \mathcal{L}_{adv} &= \min_{G_d, G_{rgb}}\max_{D} \mathbb{E}_{\mathbf{I}^r_{rgbd}}[\log D(\mathbf{I}^r_{rgbd})]  \nonumber \\
    %         &+ \mathbb{E}_{\mathbf{z}_d, \mathbf{z}_{rgb}}[\log (1-D(G_d(\mathbf{z}_d)\oplus G_{rgb}(\mathbf{z}_{rgb})))],  \label{eq:adv-loss}
\end{align}
where $\mathbf{I}^r_{rgbd}$ represents the real RGBD data, and $\mathbf{I}^f_{rgbd}$ concatenates the generated RGB image $\mathbf{I}^f_{rgb}$ and the conditioned depth $\mathbf{I}^f_{d}$.

\begin{figure}[t]
    \centering
    \includegraphics[width=1.0\linewidth]{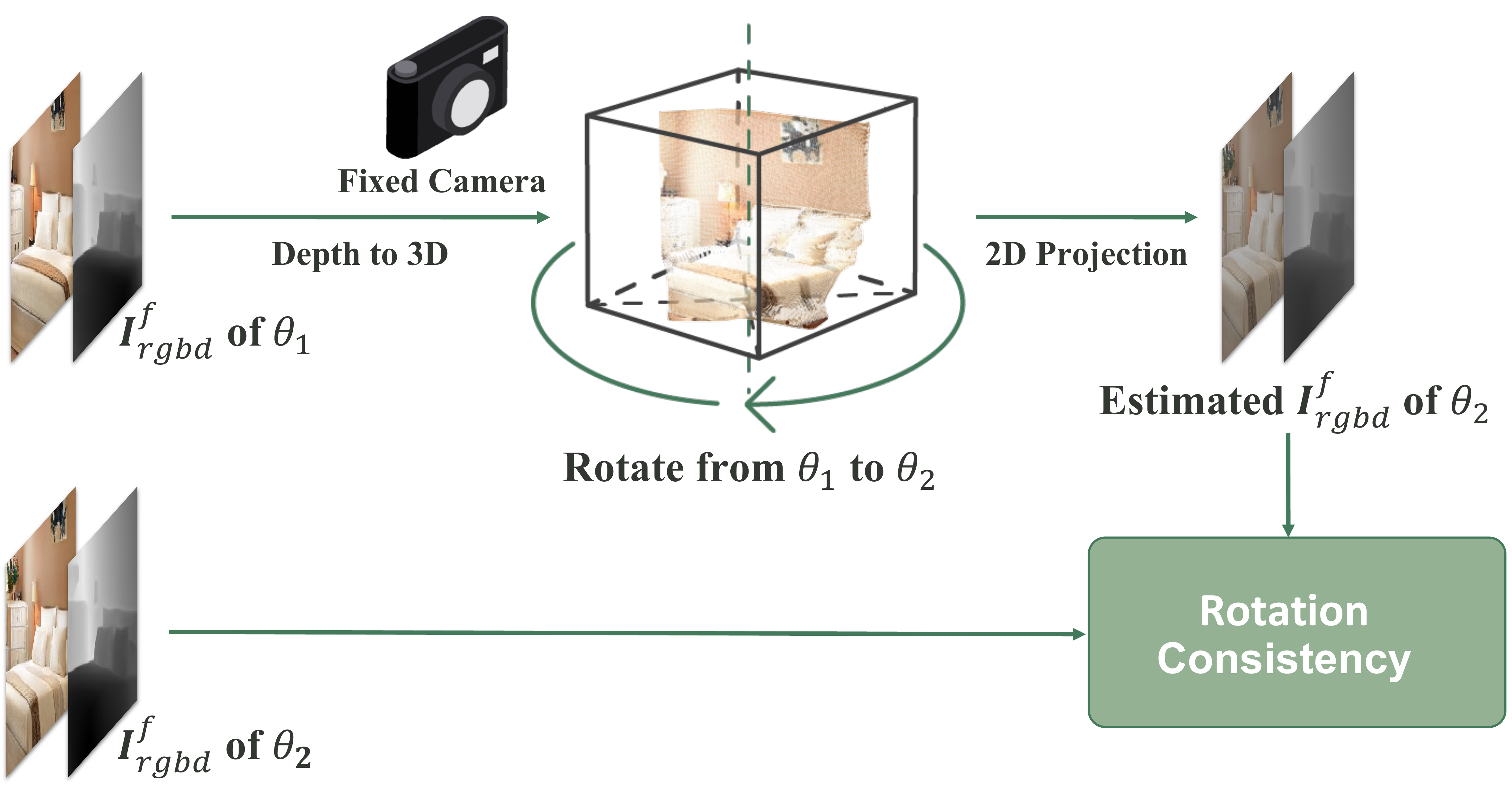}
    \vspace{-15pt}
    \caption{
        \textbf{Diagram of rotation consistency.}
        The image synthesized under angle $\theta_1$ is first projected to the 3D space as the point cloud.
        It is then rotated to $\theta_2$ around the central axis with the camera fixed.
        The rotated scene is finally re-projected to the 2D space and compared with the image generated under angle $\theta_2$.
    }
    \label{fig:Rotation}
    \vspace{-5pt}
\end{figure}

\vspace{5pt}
\noindent\textbf{Rotation Consistency Loss.}
We design the rotation consistency loss~\cite{RGBDGAN} to enhance the consistency between the synthesis from different viewpoints, \textit{i.e.}, $\theta$.
Specifically, two angles, $\theta_1$ and $\theta_2$, are randomly sampled, leading to two samples, $\mathbf{I}^f_{rgbd, 1}$ and $\mathbf{I}^f_{rgbd, 2}$ with the same latent codes, $\mathbf{z}_d$ and $\mathbf{z}_{rgb}$.
We fix the camera and rotate the scene around its central axis, as shown in \cref{fig:Rotation}.
We assume a underlying camera intrinsic parameter, $\mathbf{K}$, which is fixed in the training process.
After rotating $\mathbf{I}^f_{rgbd, 1}$ from $\theta_1$ to $\theta_2$, we will get
\begin{align}
    P(\mathbf{I}^{f, rot}_{rgbd,1}) = \mathbf{K}R(\theta_1, \theta_2)\mathbf{K}^{-1}P(\mathbf{I}^f_{rgbd,1}), \label{eq:rotation}
\end{align}
where $R(\cdot, \cdot)$ denotes the rotation operation based on the depth image, $\mathbf{I}^f_{d,1}$ and $P(\cdot)$ represents the coordinates of the pixels.
More details are available in \textit{Appendix}.

The rotation consistency losses $\mathcal{L}^d_{rot}$ and $\mathcal{L}^{rgb}_{rot}$ for the dual-path generator are then defined as
\begin{align}
    \mathcal{L}^d_{rot} &= \|\mathbf{I}^{f, rot}_{d, 1}-\mathbf{I}^f_{d, 2}\|_1, \label{eq:depth-rot} \\
    \mathcal{L}^{rgb}_{rot} &=\|\mathbf{I}^{f, rot}_{rgb, 1}-\mathbf{I}^f_{rgb, 2}\|_1, \label{eq:rgb-rot}
\end{align}
where $\|\cdot\|_1$ denotes the $\ell_1$ norm.

\vspace{5pt}
\noindent\textbf{Depth Prediction Loss.}
As discussed above, besides differentiating real and fake data, our switchable discriminator is also asked to predict the depth from a given RGB image.
Such a prediction is trained on real image-depth pairs, and further used to guide the synthesis.
Following \cite{fu2018deep}, our depth prediction is learned with a $k$-class classification.
Thus, the depth prediction branch of $D$, $D_d(\cdot)$, produces a $k$-channel output map, indicating the class probability for each pixel.
The loss function is formulated as
\begin{align}
    \mathcal{L}^r_{dp} = \mathcal{H}(D_d(\mathbf{I}^r_{rgb}), \mathbf{I}^r_d), \label{eq:depth-real}
\end{align}
where $\mathcal{H}(\cdot, \cdot)$ denotes the pixel-wise cross-entropy loss.
$\mathbf{I}^r_{rgb}$ and $\mathbf{I}^r_d$ stand for the ground-truth pair.

In order to help the generated appearance, $\mathbf{I}^f_{rgb}$, better fit the geometry, $\mathbf{I}^f_d$, we also predict the depth from the synthesized image to in turn guide the generator with
\begin{align}
    \mathcal{L}^f_{dp} = \mathcal{H}(D_d(\mathbf{I}^f_{rgb}), \mathbf{I}^f_d), \label{eq:depth-fake}
\end{align}

\vspace{5pt}
\noindent\textbf{Full Objectives.} In summary, the dual-path generator (\textit{i.e.}, $G_d$ and $G_{rgb}$) and the switchable discriminator (\textit{i.e.}, $D$) are jointly optimized with
\begin{align}
    % L = L_{adv} + \lambda_1 L^d_{rot} + \lambda_2 L^{rgb}_{rot} + \lambda_3 L^r_{dp} + \lambda_4 L^g_{dp}, \\
    \mathcal{L}_{G_d} &= \mathcal{L}^g_{adv} + \lambda_1 \mathcal{L}^d_{rot}, \\
    \mathcal{L}_{G_{rgb}} &= \mathcal{L}^g_{adv} + \lambda_2 \mathcal{L}^{rgb}_{rot} + \lambda_3 \mathcal{L}^f_{dp}, \\
    \mathcal{L}_{D} &= \mathcal{L}^d_{adv} + \lambda_4 \mathcal{L}^r_{dp},
\end{align}
where $\{\lambda_i\}_{i=1}^4$ are loss weights to balance different terms.

\begin{table*}[t]
\setlength{\tabcolsep}{8.0pt}
\centering
\caption[xxx]{
    \textbf{Quantitative comparisons} with existing 3D-aware image synthesis models on LSUN bedroom and kitchen datasets~\cite{yu2015lsun} under $128\times128$ and $256\times256$ resolutions.
    %
    % FID~\cite{fid} is used to evaluate RGB images and depths. Rotation precision (RP) and rotation consistency (RC) are used as the metrics to evaluate the synthesis quality and the 3D controllability.
    FID~\cite{fid} regarding RGB images and depths, rotation precision (RP) and rotation consistency (RC) are used as the metrics to evaluate the synthesis quality and the 3D controllability.
}

\vspace{-5pt}
\begin{tabular}{l|cccc|cccc}
\toprule
        &            \multicolumn{4}{c|}{Bedroom}           &           \multicolumn{4}{c}{Kitchen}                \\
        \hline
        &    \FID    &    \FIDD   &    \RPE    &    \RCE    &    \FID    &    \FIDD    &    \RPE    &  \RCE        \\ 
        \hline
  $ 128 \times 128 $                                                                                               \\
        \hline
  2D-GAN SeFa~\cite{shen2021closed}
        &    8.65    &    N/A     &    0.572   &   1.027    &   11.53    &     N/A     &   0.748    &  1.115        \\
%   HoloGAN~\cite{HoloGAN2019} 
%         &  210.341   &    N/A     &   0.067    &   0.630    &  230.349   &     N/A     &   0.083    &  1.221        \\
  GRAF~\cite{graf}
        &   63.940   & 184.379    &    0.218   &    1.149   &   86.920   &   239.657   &   0.224    &  1.326        \\
  GRAF-RGBD~\cite{graf}
        &   158.503  & 107.653    &    0.135   &    1.108   &   139.902  &   157.801   &   0.110    &  0.970        \\
  GIRAFFE~\cite{GIRAFFE}
        &   48.412   &  422.634   &     N/A    &     N/A    &   42.923   &   307.233   &   N/A      &  N/A          \\
  $\pi$-GAN~\cite{piGAN2021} 
        &   28.128   &  201.722   &    0.033   &    0.572   &   29.790   &   398.146   &   0.028    &  0.702        \\
  $\pi$-GAN-RGBD~\cite{piGAN2021}
        &   30.932   &  101.739   & \tb{0.022} & \tb{0.420} &   46.332   &   112.171   & \tb{0.025} & \tb{0.482}    \\
        \hline
  DepthGAN (Ours)
        % & \tb{4.040} &\tb{18.874} &   0.040    &    0.530   & \tb{5.552} & \tb{33.252} &    0.057   & 0.655         \\
        & \tb{4.040} &\tb{18.874} &   0.040    &    0.530   & \tb{5.068} & \tb{17.655} &    0.038   & 0.551         \\
        \hline
  $ 256 \times 256 $                                                                                                \\
        \hline
  2D-GAN SeFa~\cite{shen2021closed}
        &    7.19    &    N/A     &    0.401   &    1.110   &   10.85    &     N/A     &   0.480    &  1.163        \\
  GRAF~\cite{graf}
        &   66.856   &  188.368   &    0.219   &    0.880   &   94.095   &   204.050   &   0.227    &  0.928         \\
  GRAF-RGBD~\cite{graf}
        &   194.260  &  156.081   &    0.154   &    1.193   &   244.480  &   142.436   &   0.133    &  0.966         \\
  GIRAFFE~\cite{GIRAFFE}
        &   44.232   &  420.681   &     N/A    &     N/A    &   50.256   &   370.760   &    N/A     &  N/A           \\
  $\pi$-GAN~\cite{piGAN2021}
        &   48.926   &  175.744   &    0.052   &    0.597   &   41.178   &   398.946   &   0.051    &  0.726         \\
  $\pi$-GAN-RGBD~\cite{piGAN2021}
        &    49.640  &   94.196   &    0.036   &    0.510   &   77.066   &   104.865   & 0.039     &  0.566         \\
         \hline
  DepthGAN (Ours)
        % & \tb{4.797} &\tb{17.140} & \tb{0.025} & \tb{0.456} & \tb{6.168} & \tb{45.647} &    0.066   &  0.697         \\
        & \tb{4.797} &\tb{17.140} & \tb{0.025} & \tb{0.456} & \tb{6.051} & \tb{25.335} &    \tb{0.028}   &  \tb{0.502}         \\

\bottomrule
\end{tabular}
\label{tab:baselines_lsun}
\vspace{-5pt}
\end{table*}

\begin{figure*}[!ht]
    \centering
    \vspace{-5pt}
    \includegraphics[width=0.91\linewidth]{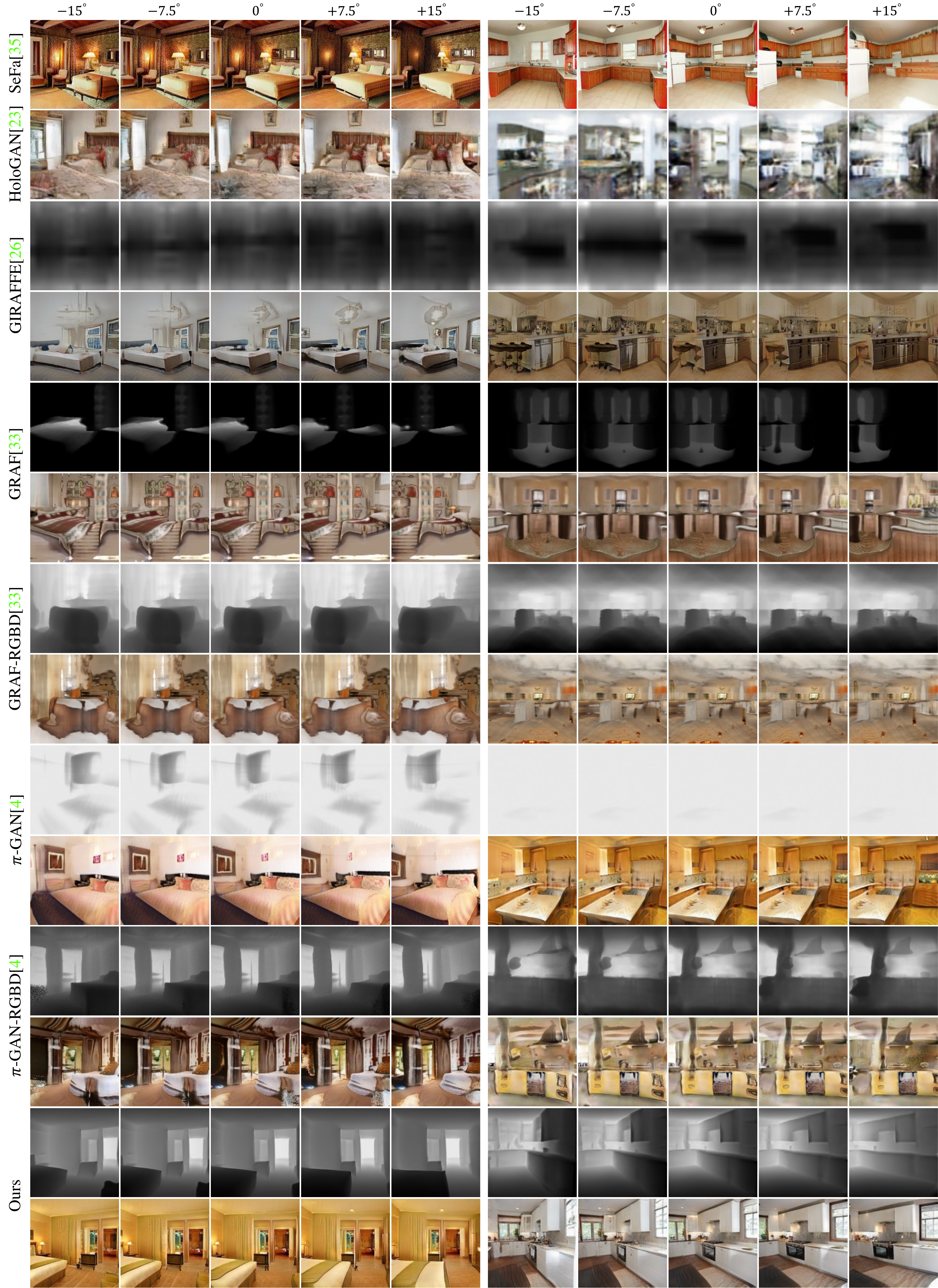}
    \vspace{-5pt}
    \caption{
        \textbf{Qualitative comparisons} with existing 3D-aware image synthesis models on LSUN bedroom and kitchen datasets~\cite{yu2015lsun}.
        Each scene is evenly rotated by 30 degrees to generate five samples.
        Zoom in for details.
    }
    \label{fig:qualitative}
    \vspace{-5pt}
\end{figure*}

\section{Experiments}\label{sec:experiment}

\noindent\textbf{Datasets.}
We conduct experiments on LSUN bedroom and kitchen datasets~\cite{yu2015lsun} to evaluate our proposed DepthGAN.
%
% For instance, the objects and layouts in the kitchen are different from image to image.
%
There are about 3M and 2M RGB images in each dataset, respectively.
Compared with the object datasets used in prior arts~\cite{graf, piGAN2021, GIRAFFE}, indoor scenes are far more challenging due to their high diversity.
We take the last 50k images from each dataset as the validation set, while the rest are used as the training set. 
In order to get the corresponding depth image of each RGB image, we utilize the pre-trained model from \cite{leres2021} to predict the depth for each image. 
The images are resized and center-cropped to the resolutions of 128 and 256 for training and evaluation.

\vspace{5pt}
\noindent\textbf{Implementation Details.}
The generators $G_d$ and $G_{rgb}$ follow the structure of StyleGAN2~\cite{stylegan2}. 
The discriminator has a similar architecture as in StyleGAN2 except for an additional channel added to the input layer and the intermediate three-level features used as input of the depth prediction network. 
We fix the focal length to 26mm. 
The angles are uniformly sampled from -15$^{\circ}$ to 15$^{\circ}$. 
The whole network is trained from scratch. More details are available in the \textit{Appendix}.

%depth shift, angle range
% We train the 

\noindent\textbf{Metrics.}
We use the following metrics to evaluate the baselines and our method: Fr\'{e}chet Inception Distance (FID) \cite{fid}, Rotation Precision (RP) and Rotation Consistency (RC). 
FID is used to evaluate the quality of both the generated RGB images and depth images. The FID evaluation for depth images is obtained by repeating the one-channel depth image to a three-channel image as input. 
In addition, we propose another two metrics for evaluation.
\textit{(1) Rotation Precision (RP)} is aimed to measure the accuracy of the angle of rotation given two generated images from two sampled angles of the same scene. The formulation is the same as \cref{eq:depth-rot}. 
\textit{(2) Rotation Consistency (RC)} targets at the rotation consistency evaluation and has the same format as \cref{eq:rgb-rot}.
Since our discriminator can be the role of depth estimator, in ablation studies, we also report the depth prediction accuracy \textit{DP (Real)} on real images from the test set. 
Besides, we predict the depth for the synthesized RGB image by the pre-trained depth prediction model\cite{leres2021}, and compare it with the generated depth to form the evaluation metric \textit{DP (Fake)}.

\noindent\textbf{Baselines.}
We compare against four state-of-the-art methods for 3D-aware image synthesis: HoloGAN~\cite{HoloGAN2019}, GRAF~\cite{graf}, GIRAFFE~\cite{GIRAFFE} and $\pi$-GAN~\cite{piGAN2021}.%
\footnote{We fail to reproduce HoloGAN with the \href{https://github.com/thunguyenphuoc/HoloGAN}{official implementation}, hence we do not report the quantitative results. The qualitative results of bedrooms are borrowed from the original paper~\cite{HoloGAN2019}, while those of kitchens are generated by our poorly reproduced model.}
For a fair comparison, we also incorporate depth information into existing methods by either employing another discriminator for depth learning or changing the input/output from RGB to RGBD image, which result in the two variants GRAF-RGBD and $\pi$-GAN-RGBD.
Another baseline is a 2D-based approach named SeFa~\cite{shen2021closed}, which can rotate the scenes through interpolation in the latent space.
% 
% All the baselines are trained from scratch. 
% 
The implementation details are available in the \textit{Appendix}. 
Since GIRAFFE~\cite{GIRAFFE} fixes the background and rotates the objects only, we do not report the RP and RC metrics on it. 
For SeFa, we use the pre-trained model \cite{leres2021} to estimate a depth map from the generated image first and then calculate the RP and RC using the predicted depth.

\subsection{Quantitative Results}
\cref{tab:baselines_lsun} reports the quantitative comparisons on LSUN bedroom and LSUN kitchen. 
We show significant improvement of image quality compared with 3D-aware image synthesis baselines in terms of the FID scores on both the RGB images and depth images.
When the 3D-aware image synthesis methods are given with the depth information for training, the quality of the generated geometry generally improves, but the quality of the appearance decreases.
While maintaining the high quality of the image, ours ensures the 3D consistency as well. Note that while $\pi$-GAN has lower RP and RC values than ours, it produces depth maps with simple geometry that is reflected by the FID on depth images and the qualitative results. It is easier to maintain consistency with simpler geometry.

\subsection{Qualitative Results}
The generated images from each baseline and our DepthGAN are shown in \cref{fig:qualitative}. 
2D-GANs can generate RGB images of high quality. However, interpolation in the latent space does not guarantee 3D consistency, and thus both the geometry and appearance can be changed during rotation.
Though 3D-aware image synthesis methods can synthesize RGB images of discernible scenes, they generally fail to learn a reasonable geometry unsupervisedly for both the bedrooms and the kitchens. 
This indicates that in the previous works, generating a visually-pleasing RGB image does not require a good understanding of the underlying 3D geometry. 
Besides, the image quality degrades significantly compared with that of 2D GANs. 
With the help of ground-truth depth information, GRAF-RGBD and $\pi$-GAN-RGBD can generate geometries of higher quality but sacrifice the quality of the appearance. 
In contrast, our DepthGAN can generate images with reasonable geometries and photo-realistic appearance simultaneously, which mitigates the gap between the 3D-aware image synthesis and 2D GANs and surpass the recent 3D-aware image synthesis methods on scene generation as well.
%
% Noticeably, $\pi$-GAN ...

\begin{table*}[t]
\setlength{\tabcolsep}{12.5pt}
\centering 
\caption{
    \textbf{Ablation study} conducted on LSUN bedroom dataset~\cite{yu2015lsun} under $128\times128$ resolution.
    FID~\cite{fid} regarding RGB images and depths, rotation precision (RP) and rotation consistency (RC), depth prediction (DP) on real and fake samples are used as the metrics.
}
\vspace{-8pt}
\begin{tabular}{l|cccccc}
           \toprule
           &     \FID   &    \FIDD    &    \RPE    &    \RCE    &   \RealD   &   \FakeD  \\ 
           \hline
  \textit{w/o} $L_{dp}^r$ \& $L_{dp}^f$  
           &    4.882   &     26.518  &    0.067   &   0.711    &     N/A    &  0.317     \\
  \textit{w/o} $L_{dp}^f$ 
        %   &    5.484   &     27.161  &    0.061   &   0.705    &    1.337   &   0.318   \\
            &    5.441   &     24.633  &    0.066   &   0.683    &    1.334   &   0.318   \\
  \textit{w/o} $L_{rgb}^{rot}$
            &    5.038   &     24.834  &    0.067   &   0.716    &    1.303   &   0.315   \\
  \textit{w/o} $L_{d}^{rot}$
        %   &    6.447   & \tb{15.111} &    0.152   &   1.176    &    1.337   &   0.311   \\
            &    4.504   & \tb{8.315} &    0.152   &   1.196    &    1.279   &   0.311   \\
  \textit{w/o} condition
           &    24.062   &  119.917   &   0.097    & 1.443      &    1.242   & 0.343  \\
  \textit{w/o} condition \& rotation
           &    \tb{2.793}   &     21.225  &     N/A    &    N/A     &    1.205   &   0.312   \\
     Ours-full 
           & 4.040 &    18.874   & \tb{0.040} & \tb{0.530} & \tb{1.201} & \tb{0.310}  \\
% Ours& 4.797 & 0.023 &0.539 &1.217 & 0.177 256

\bottomrule
\end{tabular}
\label{tab:ablations} 
\vspace{-5pt}
\end{table*}

\subsection{Ablation Study}
We analyze the effectiveness of each component of the DepthGAN. Evaluation results are shown in the \cref{tab:ablations}.
There is a discrepancy between the generated depth and RGB images if there is no depth prediction loss. As such, the discriminator struggles to lead the generator to capture a coherent relationship between the depth and appearance, and all the metrics can be observed with obvious drops.
Without the rotation-consistency loss on RGB images, the RGB consistency completely depends on the conditioning and the discriminator, which forces the network to figure out the consistency on RGB images by itself. While the network is working hard to learn such consistency, it also hinders other aspects of learning to some extent.
To test the performance of DepthGAN without rotation-consistency loss on depth, we allow the rotation-consistency loss on RGB images to backpropagate the gradients to $G_{d}$, which is different from our original design. Without the consistency loss on the rotation of depth images, there are fewer constraints on the depth generation, and thus this results in a lower FID score on the generated depth images. However, the rotation precision and consistency measurements experience a significant drop due to the lack of explicit supervision on the depth rotation. 
We also report the result without conditioning appearance features on depth features. For discriminator, the depth prediction from a real image is preserved to enhance the 3D knowledge within it. When rotation consistency loss is included for training, where the generator has the same structure as RGBD-GAN~\cite{RGBDGAN}, the network is unable to capture the correct depth-appearance pair. If rotation consistency loss is removed, where the generator is the same with that of StyleGAN2~\cite{stylegan2}, the network fails to view the scene from different angles directly and thus lacks 3D knowledge although the FID score on RGB images is lower.
% We also report the result without conditioning appearance features on depth features, where the generator has the same structure as StyleGAN2. For discriminator, the depth prediction from a real image is kept to enhance the 3D knowledge within it. Though the FID score on RGB images is lower, it fails to view the scene from different angles directly and thus lacks 3D knowledge.

\begin{figure}[t]
    \centering
    \includegraphics[width=1.0\linewidth]{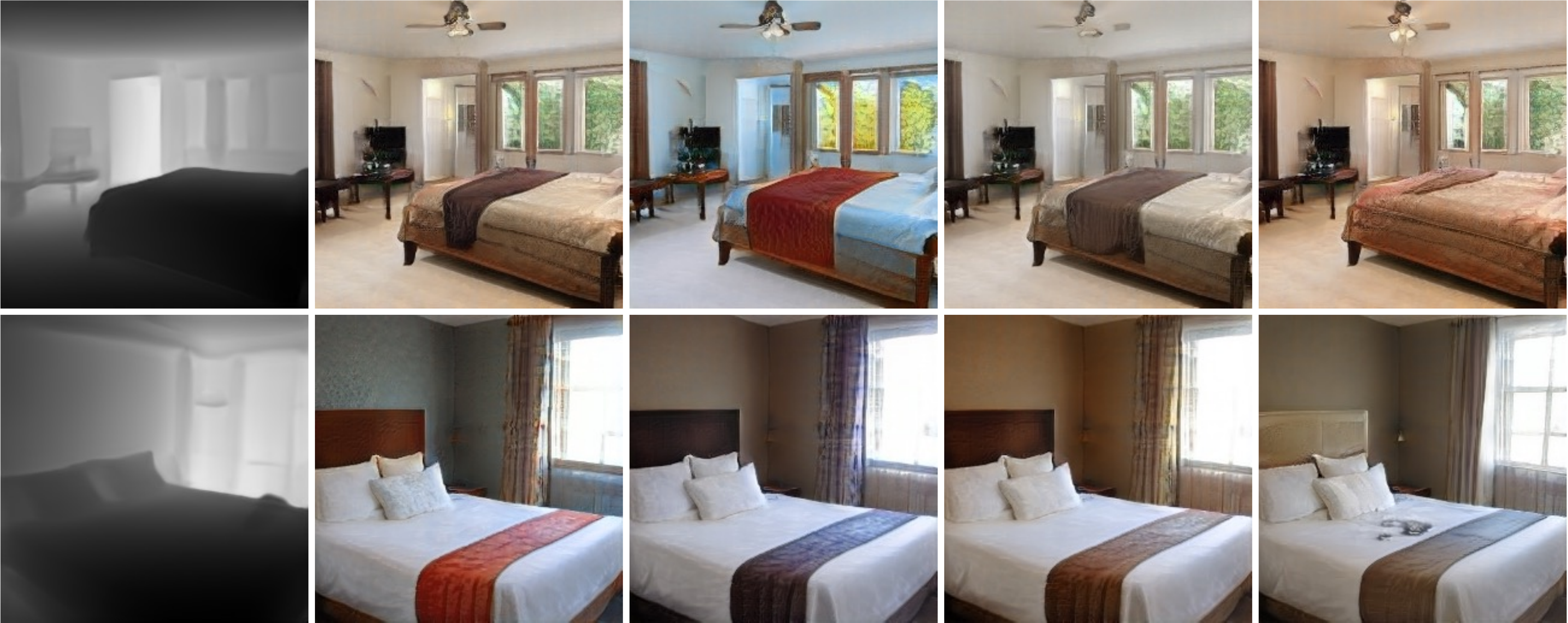}
    \vspace{-15pt}
    \caption{
        \textbf{Diverse synthesis via varying the appearance latent code $\mathbf{z}_{rgb}$}, conditioned on the same depth image.
    }
    \label{fig:depthfixed}
    % \vspace{-5pt}
\end{figure}

\begin{figure}[t]
    \centering
    \includegraphics[width=1.0\linewidth]{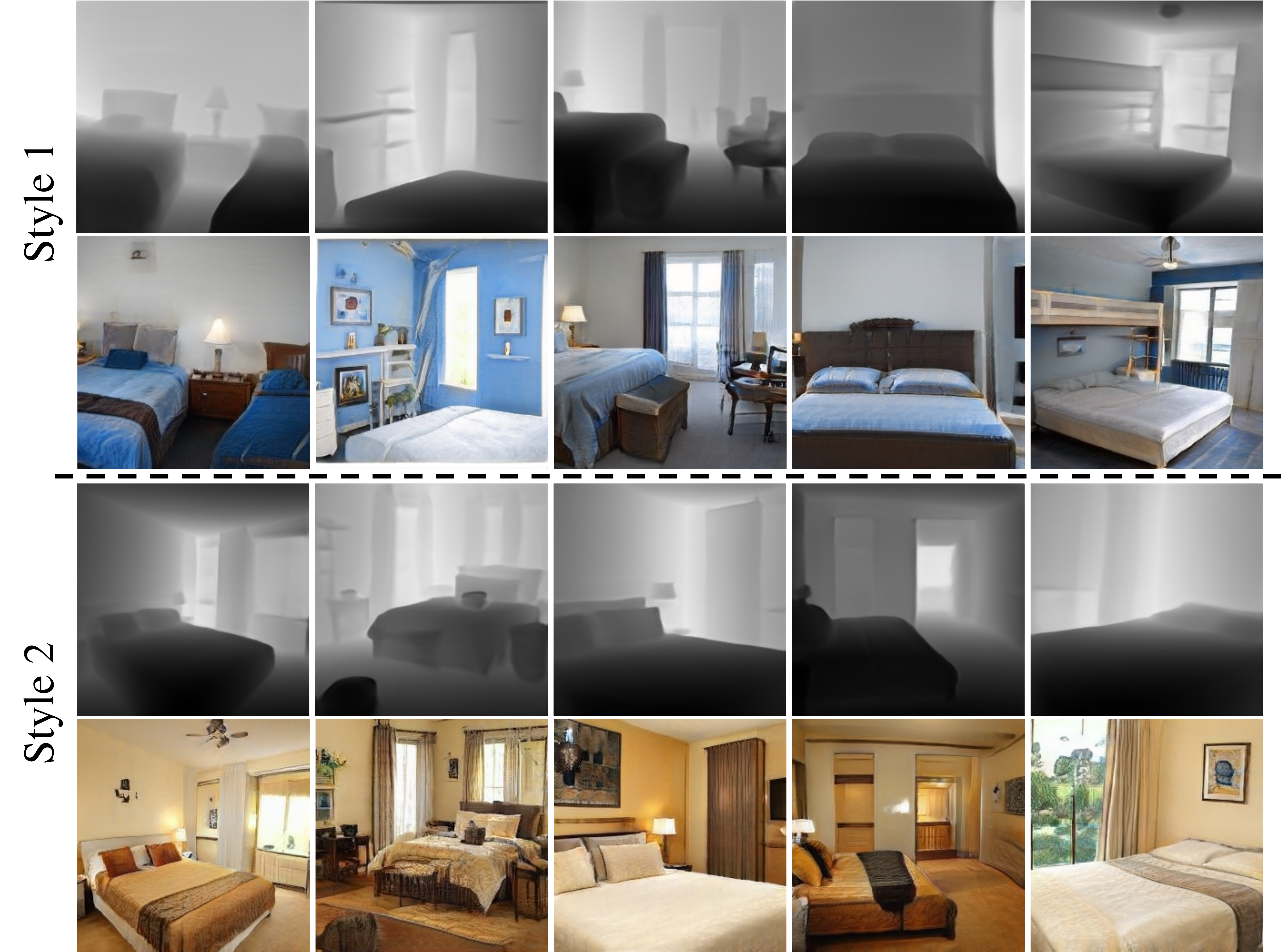}
    \vspace{-15pt}
    \caption{
        \textbf{Diverse geometries via varying the depth latent code $\mathbf{z}_d$}, rendered with the same appearance style.
    }
    \label{fig:rgbfixed}
    \vspace{-5pt}
\end{figure}

\begin{figure*}[t]
    \centering
    \includegraphics[width=1.0\linewidth]{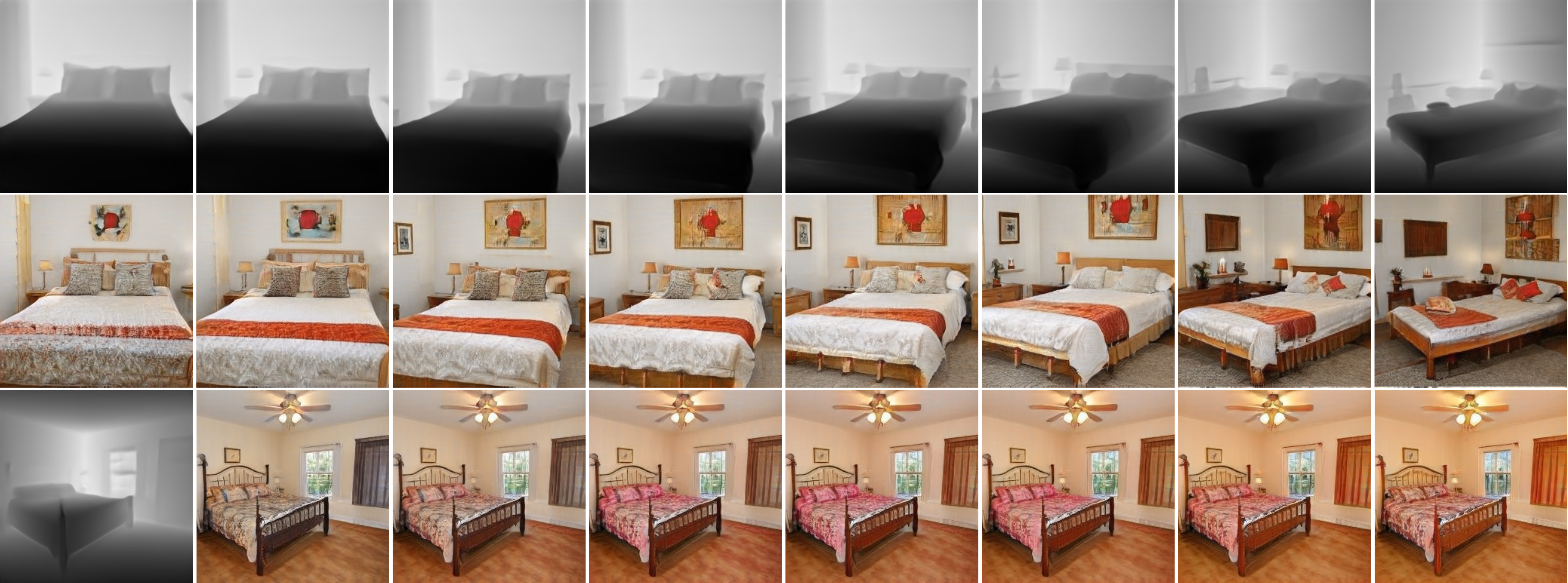}
    \vspace{-20pt}
    \caption{
        \textbf{Interpolation results} regarding depth (\textit{i.e.}, the first two rows) and appearance (\textit{i.e.}, the bottom row).
        Note that interpolation in the latent space is \textit{different} from rotation as the 3D consistency is not guaranteed.
    }
    \label{fig:interpolation}
    \vspace{-10pt}
\end{figure*}

\subsection{Controllable Image Synthesis}
\noindent\textbf{Disentanglement.} With the design of the dual-path generator, the latent spaces of the two generators are separate and thus can be sampled independently. This allows for clear disentanglement of the geometry and appearance. 
\cref{fig:depthfixed} shows the cases where the latent codes for depth generation are fixed, and the latent codes are changed for varying appearance. 
The underlying 3D geometries are the same for all the images within the same row while the styles keep changing.
On the contrary, images in \cref{fig:rgbfixed} share the same style but the geometries are various, which is brought by fixed latent code for $G_{rgb}$ and multiple latent codes for $G_{d}$. 

\noindent\textbf{Linearity.} To demonstrate that two latent spaces learned by DepthGAN are semantically meaningful, we linearly interpolate between two latent codes from one latent space and fix the latent code from the other latent space. The interpolation results are shown in \cref{fig:interpolation}. 
Note that though the visual effect from the interpolation in the depth latent space is similar to that of 3D-aware rotation, they are different as there is no guarantee for 3D consistency during interpolation.

\subsection{Discussion}
% While DepthGAN has the capability to synthesize photo-realistic multi-view-consistent images, there are several limitations to be solved in the future work. 
 
\noindent\textbf{Rotation.} Although the choice of rotation axis as the central one relieves the constraint that the camera stays in the same sphere with all scenes located on the center, it brings a large variety of rotation distributions. Thus, during the rotation of a generated scene, the newly generated view may be out of the manifold learned during training and render unsatisfactory images. The access to the prior distribution of the rotation axes from real data may ease the problem. As the current angle range for rotation is from $-15^\circ$ to $15^\circ$, we do not take special treatment for occlusion. However, if the angle range is required to be larger, the occlusion is an inevitable problem to be handled, which we leave for future exploration.

\noindent\textbf{Ground-Truth 3D Information.} The quality of the generated 3D-aware images highly relies on the performance of the pre-trained depth prediction methods. We notice that for some objects such as light on the ceiling and some windows or paintings on the wall, there is no depth information available (e.g., images in \cref{fig:depthfixed} and \cref{fig:rgbfixed}). This is due to the fact that the pre-trained depth prediction model fails to predict depths for such minute details, and the generator tends to generate the geometry from the distribution learned from the ground-truth depth images. Introducing real depth images collected by machines into the training should alleviate this limitation.

\begin{figure}[t]
    \centering
    \includegraphics[width=1.0\linewidth]{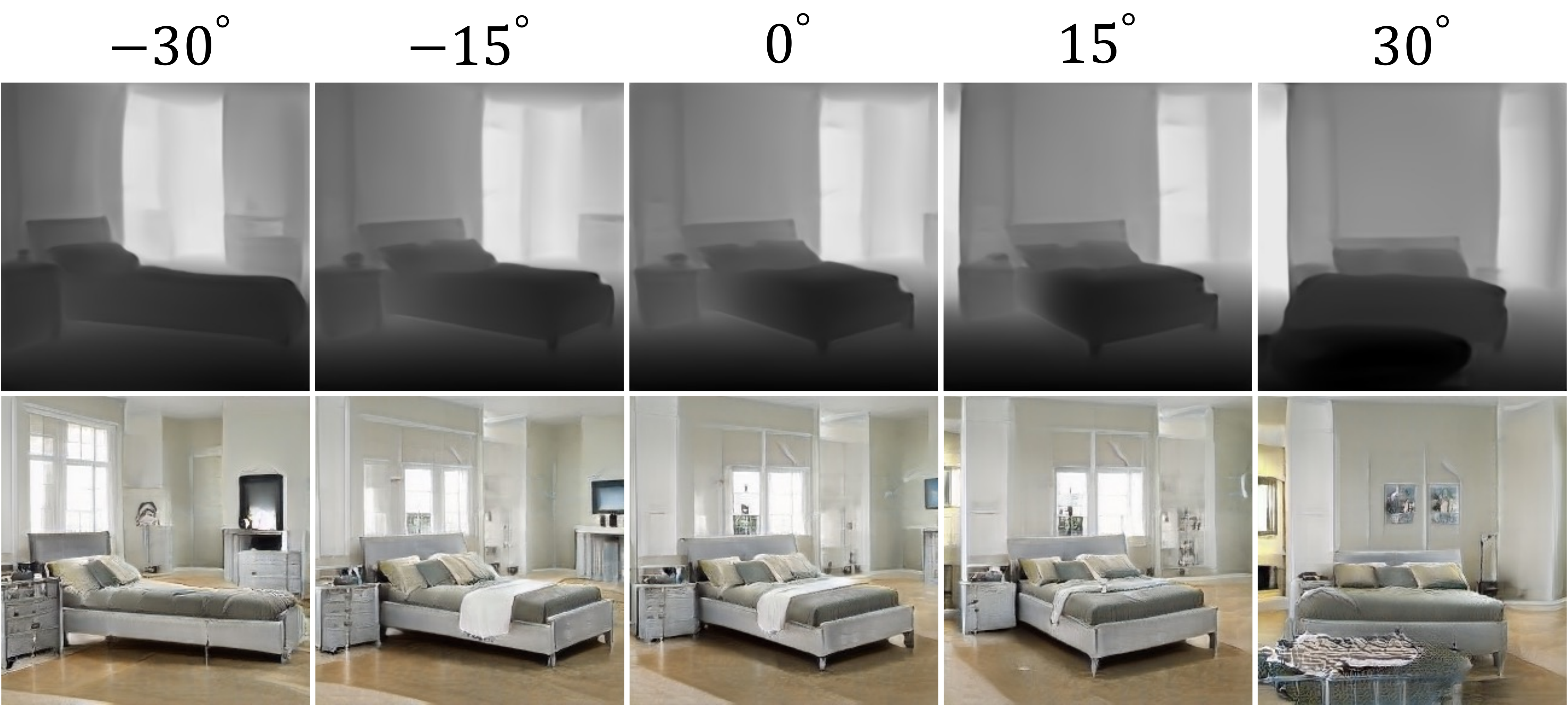}
    \vspace{-20pt}
    \caption{
        \textbf{Out-of-distribution generation.}
    }
    \label{fig:out_of_dist}
    \vspace{-15pt}
\end{figure}

\noindent\textbf{Out-of-Distribution Generation.} During training, we sample the angle of rotation from $-15^\circ$ to $15^\circ$. When extrapolating the angle outside of that range, as shown in \cref{fig:out_of_dist}, the model can generate the rotated geometry but lacks the 3D consistency as expected.

% 1. Rotation axis
% 2. Occlusion
% 3. light on the ceiling, window/painting on the wall (no depth)

% out of distribution

% pre-trained depth prediction methods
\section{Conclusion}\label{sec:conclusion}

In this work, we present DepthGAN, which can learn the appearance and the underlying geometry of indoor scenes simultaneously. 
A dual-path generator and a switchable discriminator are carefully designed to make sufficient use of the depth prior.
%
% DepthGAN takes depth as the 3D prior to facilitate the learning of 3D-aware generation.
%
% The dual-path generator and switchable discriminator are carefully designed to ensure the 3D consistency of the generated multi-view images.
% 
Experimental results demonstrate the superiority of our approach over existing methods from both the image quality and the 3D controllability perspectives.

%%%% References
{\small
\bibliographystyle{ieee_fullname}
\bibliography{ref}

\begin{thebibliography}{10}\itemsep=-1pt

\bibitem{2018GIS}
Hassan~Abu Alhaija, Siva~Karthik Mustikovela, Andreas Geiger, and Carsten
  Rother.
\newblock Geometric image synthesis.
\newblock In {\em Asian Conf. Comput. Vis.}, 2018.

\bibitem{ashburner2000voxel}
John Ashburner and Karl~J Friston.
\newblock Voxel-based morphometry—the methods.
\newblock {\em Neuroimage}, 2000.

\bibitem{biggan}
Andrew Brock, Jeff Donahue, and Karen Simonyan.
\newblock Large scale {GAN} training for high fidelity natural image synthesis.
\newblock In {\em Int. Conf. Learn. Represent.}, 2019.

\bibitem{piGAN2021}
Eric Chan, Marco Monteiro, Petr Kellnhofer, Jiajun Wu, and Gordon Wetzstein.
\newblock pi-{GAN}: Periodic implicit generative adversarial networks for
  3d-aware image synthesis.
\newblock In {\em IEEE Conf. Comput. Vis. Pattern Recog.}, 2021.

\bibitem{chen2021ngp}
Xuelin Chen, Daniel Cohen-Or, Baoquan Chen, and Niloy~J. Mitra.
\newblock Towards a neural graphics pipeline for controllable image generation.
\newblock {\em Computer Graphics Forum}, 2021.

\bibitem{cheung2000real}
German~KM Cheung, Takeo Kanade, J-Y Bouguet, and Mark Holler.
\newblock A real time system for robust 3d voxel reconstruction of human
  motions.
\newblock In {\em IEEE Conf. Comput. Vis. Pattern Recog.}, 2000.

\bibitem{cityscapes}
Marius Cordts, Mohamed Omran, Sebastian Ramos, Timo Rehfeld, Markus Enzweiler,
  Rodrigo Benenson, Uwe Franke, Stefan Roth, and Bernt Schiele.
\newblock The cityscapes dataset for semantic urban scene understanding.
\newblock In {\em IEEE Conf. Comput. Vis. Pattern Recog.}, 2016.

\bibitem{GSN}
Terrance DeVries, Miguel~Angel Bautista, Nitish Srivastava, Graham~W. Taylor,
  and Joshua~M. Susskind.
\newblock Unconstrained scene generation with locally conditioned radiance
  fields.
\newblock In {\em Int. Conf. Comput. Vis.}, 2021.

\bibitem{fu2018deep}
Huan Fu, Mingming Gong, Chaohui Wang, Kayhan Batmanghelich, and Dacheng Tao.
\newblock Deep ordinal regression network for monocular depth estimation.
\newblock In {\em IEEE Conf. Comput. Vis. Pattern Recog.}, 2018.

\bibitem{gan}
Ian Goodfellow, Jean Pouget-Abadie, Mehdi Mirza, Bing Xu, David Warde-Farley,
  Sherjil Ozair, Aaron Courville, and Yoshua Bengio.
\newblock Generative adversarial nets.
\newblock In {\em Adv. Neural Inform. Process. Syst.}, 2014.

\bibitem{gu2021stylenerf}
Jiatao Gu, Lingjie Liu, Peng Wang, and Christian Theobalt.
\newblock Stylenerf: A style-based 3d-aware generator for high-resolution image
  synthesis.
\newblock {\em arXiv preprint arXiv:2110.08985}, 2021.

\bibitem{ganspace}
Erik H{\"a}rk{\"o}nen, Aaron Hertzmann, Jaakko Lehtinen, and Sylvain Paris.
\newblock {GANSpace}: Discovering interpretable {GAN} controls.
\newblock In {\em Adv. Neural Inform. Process. Syst.}, 2020.

\bibitem{fid}
Martin Heusel, Hubert Ramsauer, Thomas Unterthiner, Bernhard Nessler, and Sepp
  Hochreiter.
\newblock {GANs} trained by a two time-scale update rule converge to a local
  nash equilibrium.
\newblock In {\em Adv. Neural Inform. Process. Syst.}, 2017.

\bibitem{pggan}
Tero Karras, Timo Aila, Samuli Laine, and Jaakko Lehtinen.
\newblock Progressive growing of {GAN}s for improved quality, stability, and
  variation.
\newblock In {\em Int. Conf. Learn. Represent.}, 2018.

\bibitem{stylegan3}
Tero Karras, Miika Aittala, Samuli Laine, Erik H\"ark\"onen, Janne Hellsten,
  Jaakko Lehtinen, and Timo Aila.
\newblock Alias-free generative adversarial networks.
\newblock In {\em Adv. Neural Inform. Process. Syst.}, 2021.

\bibitem{stylegan}
Tero Karras, Samuli Laine, and Timo Aila.
\newblock A style-based generator architecture for generative adversarial
  networks.
\newblock In {\em IEEE Conf. Comput. Vis. Pattern Recog.}, 2019.

\bibitem{stylegan2}
Tero Karras, Samuli Laine, Miika Aittala, Janne Hellsten, Jaakko Lehtinen, and
  Timo Aila.
\newblock Analyzing and improving the image quality of {StyleGAN}.
\newblock In {\em IEEE Conf. Comput. Vis. Pattern Recog.}, 2020.

\bibitem{kitti}
Moritz Menze and Andreas Geiger.
\newblock Object scene flow for autonomous vehicles.
\newblock In {\em IEEE Conf. Comput. Vis. Pattern Recog.}, 2015.

\bibitem{which2018}
Lars Mescheder, Sebastian Nowozin, and Andreas Geiger.
\newblock Which training methods for gans do actually converge?
\newblock In {\em Int. Conf. Mach. Learn.}, 2018.

\bibitem{michalkiewicz2019implicit}
Mateusz Michalkiewicz, Jhony~K Pontes, Dominic Jack, Mahsa Baktashmotlagh, and
  Anders Eriksson.
\newblock Implicit surface representations as layers in neural networks.
\newblock In {\em Int. Conf. Comput. Vis.}, 2019.

\bibitem{nerf}
Ben Mildenhall, Pratul~P Srinivasan, Matthew Tancik, Jonathan~T Barron, Ravi
  Ramamoorthi, and Ren Ng.
\newblock Nerf: Representing scenes as neural radiance fields for view
  synthesis.
\newblock In {\em Eur. Conf. Comput. Vis.}, 2020.

\bibitem{nyu}
Pushmeet~Kohli Nathan~Silberman, Derek~Hoiem and Rob Fergus.
\newblock Indoor segmentation and support inference from rgbd images.
\newblock In {\em Eur. Conf. Comput. Vis.}, 2012.

\bibitem{HoloGAN2019}
Thu Nguyen-Phuoc, Chuan Li, Lucas Theis, Christian Richardt, and Yong-Liang
  Yang.
\newblock {HoloGAN}: Unsupervised learning of 3d representations from natural
  images.
\newblock In {\em Int. Conf. Comput. Vis.}, 2019.

\bibitem{BlockGAN2020}
Thu Nguyen-Phuoc, Christian Richardt, Long Mai, Yong-Liang Yang, and Niloy
  Mitra.
\newblock {BlockGAN}: Learning 3d object-aware scene representations from
  unlabelled images.
\newblock In {\em Adv. Neural Inform. Process. Syst.}, Nov 2020.

\bibitem{campari}
Michael Niemeyer and Andreas Geiger.
\newblock Campari: Camera-aware decomposed generative neural radiance fields.
\newblock {\em arXiv preprint arXiv:2103.17269}, 2021.

\bibitem{GIRAFFE}
Michael Niemeyer and Andreas Geiger.
\newblock {GIRAFFE}: Representing scenes as compositional generative neural
  feature fields.
\newblock In {\em IEEE Conf. Comput. Vis. Pattern Recog.}, 2021.

\bibitem{RGBDGAN}
Atsuhiro Noguchi and Tatsuya Harada.
\newblock {RGBD-GAN}: Unsupervised 3d representation learning from natural
  image datasets via rgbd image synthesis.
\newblock In {\em Int. Conf. Learn. Represent.}, 2020.

\bibitem{ohtake2004ridge}
Yutaka Ohtake, Alexander Belyaev, and Hans-Peter Seidel.
\newblock Ridge-valley lines on meshes via implicit surface fitting.
\newblock In {\em ACM SIGGRAPH}, 2004.

\bibitem{midas}
Ren\'{e} Ranftl, Katrin Lasinger, David Hafner, Konrad Schindler, and Vladlen
  Koltun.
\newblock Towards robust monocular depth estimation: Mixing datasets for
  zero-shot cross-dataset transfer.
\newblock {\em IEEE Trans. Pattern Anal. Mach. Intell.}, 2020.

\bibitem{rusinkiewicz2001efficient}
Szymon Rusinkiewicz and Marc Levoy.
\newblock Efficient variants of the icp algorithm.
\newblock In {\em Int. Conf. on 3-D digital imaging and modeling}, 2001.

\bibitem{rusu20113d}
Radu~Bogdan Rusu and Steve Cousins.
\newblock 3d is here: Point cloud library (pcl).
\newblock In {\em IEEE Int. Conf. on Robotics and Automation}, 2011.

\bibitem{schnabel2007efficient}
Ruwen Schnabel, Roland Wahl, and Reinhard Klein.
\newblock Efficient ransac for point-cloud shape detection.
\newblock In {\em Comput. Graph. Forum}, 2007.

\bibitem{graf}
Katja Schwarz, Yiyi Liao, Michael Niemeyer, and Andreas Geiger.
\newblock {GRAF}: Generative radiance fields for 3d-aware image synthesis.
\newblock In {\em Adv. Neural Inform. Process. Syst.}, 2020.

\bibitem{shen2020interfacegan}
Yujun Shen, Ceyuan Yang, Xiaoou Tang, and Bolei Zhou.
\newblock {InterFaceGAN}: Interpreting the disentangled face representation
  learned by {GANs}.
\newblock {\em IEEE Trans. Pattern Anal. Mach. Intell.}, 2020.

\bibitem{shen2021closed}
Yujun Shen and Bolei Zhou.
\newblock Closed-form factorization of latent semantics in {GANs}.
\newblock In {\em IEEE Conf. Comput. Vis. Pattern Recog.}, 2021.

\bibitem{sitzmann2019siren}
Vincen Sitzmann, Julien~N.P. Martel, Alexander~W. Bergman, David~B. Lindell,
  and Gordon Wetzstein.
\newblock Implicit neural representation with periodic activation functions.
\newblock In {\em Adv. Neural Inform. Process. Syst.}, 2020.

\bibitem{replica}
Julian Straub, Thomas Whelan, Lingni Ma, Yufan Chen, Erik Wijmans, Simon Green,
  Jakob~J. Engel, Raul Mur-Artal, Carl Ren, Shobhit Verma, Anton Clarkson,
  Mingfei Yan, Brian Budge, Yajie Yan, Xiaqing Pan, June Yon, Yuyang Zou,
  Kimberly Leon, Nigel Carter, Jesus Briales, Tyler Gillingham, Elias Mueggler,
  Luis Pesqueira, Manolis Savva, Dhruv Batra, Hauke~M. Strasdat, Renzo~De
  Nardi, Michael Goesele, Steven Lovegrove, and Richard Newcombe.
\newblock The {R}eplica dataset: A digital replica of indoor spaces.
\newblock {\em arXiv preprint arXiv:1906.05797}, 2019.

\bibitem{SSGAN2016}
Xiaolong Wang and Abhinav Gupta.
\newblock Generative image modeling using style and structure adversarial
  networks.
\newblock In {\em Eur. Conf. Comput. Vis.}, 2016.

\bibitem{3dgan}
Jiajun Wu, Chengkai Zhang, Tianfan Xue, William~T Freeman, and Joshua~B
  Tenenbaum.
\newblock Learning a probabilistic latent space of object shapes via 3{D}
  generative-adversarial modeling.
\newblock In {\em Adv. Neural Inform. Process. Syst.}, 2016.

\bibitem{higan}
Ceyuan Yang, Yujun Shen, and Bolei Zhou.
\newblock Semantic hierarchy emerges in deep generative representations for
  scene synthesis.
\newblock {\em Int. J. Comput. Vis.}, 2020.

\bibitem{leres2021}
Wei Yin, Jianming Zhang, Oliver Wang, Simon Niklaus, Long Mai, Simon Chen, and
  Chunhua Shen.
\newblock Learning to recover 3d scene shape from a single image.
\newblock In {\em IEEE Conf. Comput. Vis. Pattern Recog.}, 2021.

\bibitem{yu2015lsun}
Fisher Yu, Ari Seff, Yinda Zhang, Shuran Song, Thomas Funkhouser, and Jianxiong
  Xiao.
\newblock {LSUN}: Construction of a large-scale image dataset using deep
  learning with humans in the loop.
\newblock {\em arXiv preprint arXiv:1506.03365}, 2015.

\bibitem{Zhou2018}
Qian-Yi Zhou, Jaesik Park, and Vladlen Koltun.
\newblock {Open3D}: {A} modern library for {3D} data processing.
\newblock {\em arXiv:1801.09847}, 2018.

\bibitem{peng2021}
Jiapeng Zhu, Ruili Feng, Yujun Shen, Deli Zhao, Zhengjun Zha, Jingren Zhou, and
  Qifeng Chen.
\newblock Low-rank subspaces in gans.
\newblock In {\em Adv. Neural Inform. Process. Syst.}, 2021.

\bibitem{VON}
Jun-Yan Zhu, Zhoutong Zhang, Chengkai Zhang, Jiajun Wu, Antonio Torralba,
  Joshua~B. Tenenbaum, and William~T. Freeman.
\newblock Visual object networks: Image generation with disentangled 3{D}
  representations.
\newblock In {\em Adv. Neural Inform. Process. Syst.}, 2018.

\end{thebibliography}
}

\newpage
\section*{Appendix}
\setcounter{section}{0}
\renewcommand\thesection{\Alph{section}}
\section{Implementation Details}

\subsection{Network Architecture}

\noindent\textbf{Dual-path Generator.} We use the generator of StyleGAN2~\cite{stylegan2} as the backbone of the depth generator and the appearance renderer. The output of depth generator is the one-channel depth image while the output of appearance renderer is the three-channel RGB image. The path length regularization for both generators is removed.

\begin{figure}[t]
    \centering
    \includegraphics[width=0.95\linewidth]{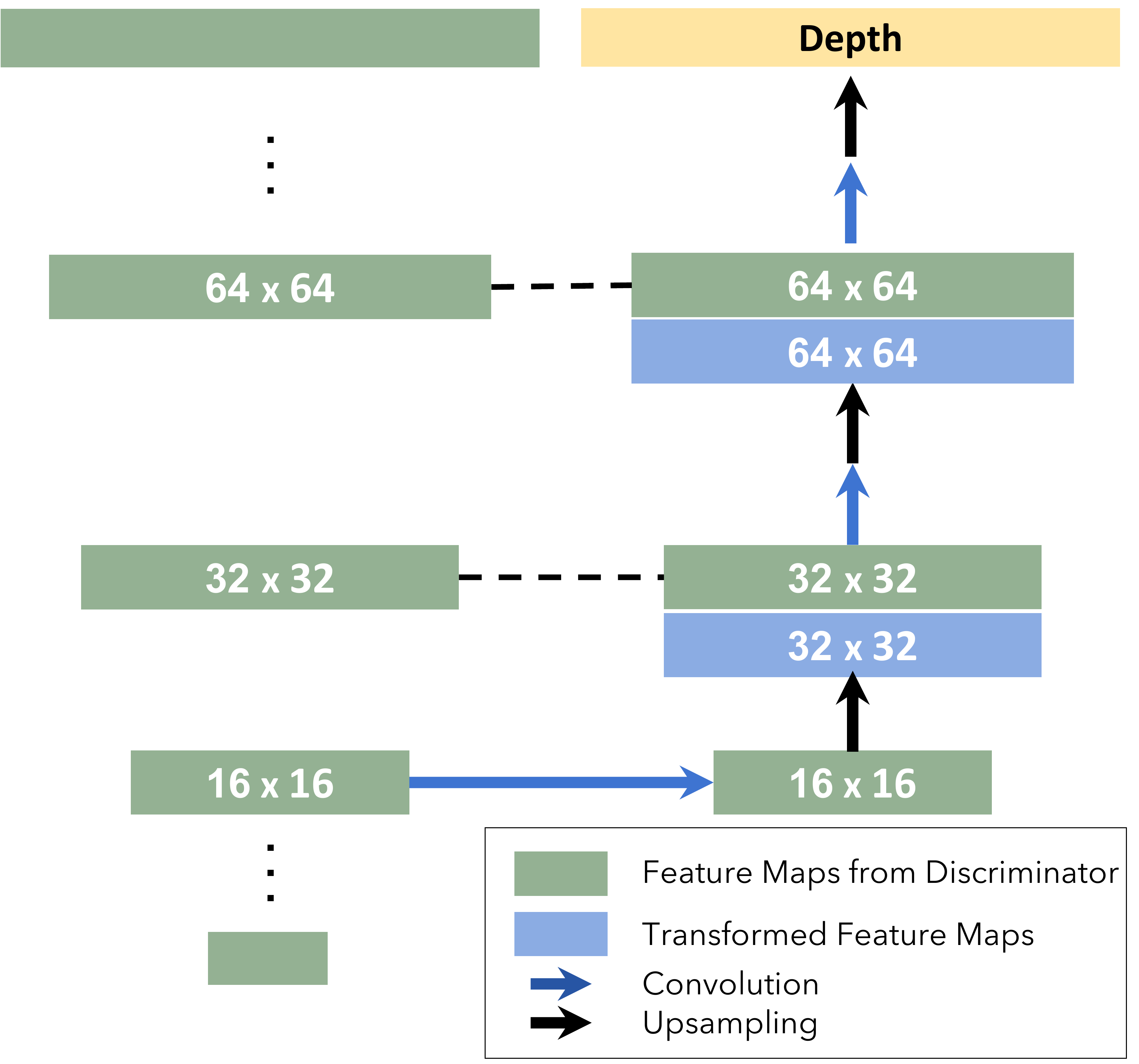}
    \caption{
        Architecture of the depth prediction network.
    }
    \label{fig:depth_pred}
    \vspace{-15pt}
\end{figure}

\noindent\textbf{Switchable Discriminator.} The switchable discriminator follows the structure of the discriminator in StyleGAN2~\cite{stylegan2}. We implement the switchable input layer $\mathbf{\Gamma}_1$ by adding one more layer for the depth input, where the input channel size is 1 and the output has the same dimension as that of the original input layer $\mathbf{\Gamma}_0$. The original input layer is dedicated for the RGB input. If the input of the discriminator is an RGB image, only $\mathbf{\Gamma}_0$ will be activated and $\mathbf{\Gamma}_1$ is disabled. If the input is an RGBD image, the RGB image goes through $\mathbf{\Gamma}_0$ and the depth image goes through $\mathbf{\Gamma}_1$. The output feature maps will be added together and be forwarded to the next layer. The architecture of depth prediction network is shown in \cref{fig:depth_pred}. It uses the features at resolution $16^2$, $32^2$, and $64^2$ from the discriminator as input. Starting from the resolution $16^2$, (1) the feature maps with smaller resolution will be fed into an one-layer convolution and activated by leaky-ReLU. (2) Then, the transformed feature maps will go through a two-layer convolution with residual connection to further enhance the useful features. (3) After that, the output features are upsampled to larger resolution and concatenated with the features from the discriminator at larger resolution. (4) The concatenated features will repeat the process starting from (1). Since we form the depth prediction as a $k$-class classification problem, the final layer output a k-channel map which indicates the class probability of each pixel.

\subsection{Training Protocol}
All the components of our proposed DepthGAN are trained in turn. When only one part is trained, the gradients of other parts will be turned off. 
% The full training process is available in \cref{alg:algo}.
% 
Our training process is as follows: (1) The dual-path generator is updated with adversarial loss $\mathcal{L}_{adv}^{g}$. (2) The depth generator is updated with rotation consistency loss $\mathcal{L}_{rot}^{d}$. (3) The appearance renderer is updated with rotation consistency loss $\mathcal{L}_{rot}^{rgb}$ and depth prediction loss $\mathcal{L}_{dp}^{f}$. (4) The discriminator is updated with adversarial loss $\mathcal{L}_{adv}^{d}$, depth prediction loss $\mathcal{L}_{dp}^r$ and $R_1$ regularization~\cite{which2018}. 
We train with Adam optimizer and use a batch size of 64. The learning rate for both the generator and the discriminator is 1.5e-3. The weight of the R1 regularization is 0.3 for resolution $128^2$ and 0.5 for resolution $256^2$. $\{\lambda_i\}_{i=1}^4$ are set to 50, 0.3, 1e-3 and 0.8 at resolution $128^2$. At resolution $256^2$, $\{\lambda_i\}_{i=1, i\neq2}^4$ are set to 50, 0.001, and 0.8, while $\lambda_2$ is 0.5 for LSUN bedroom and 0.4 for LSUN kitchen. All the experiments run on 8 Tesla V100 GPUs for about 2-3 days.

\section{Baselines}
We ensure that all the training of baselines is converged. The field of views are the same as ours. When testing, the range of azimuth is set to 30 degrees and the elevation is fixed to the frontal view.  The details for each baseline are as follows:

\noindent\textbf{SeFa}~\cite{shen2021closed}.
We factorize the weights in the first three layers of StyleGAN2~\cite{stylegan2} and find the most relevant direction that affects the pose of the bedroom or kitchen.
The pre-trained models can be found in \href{https://github.com/genforce/genforce}{GenForce}.
When editing the pose, the truncation is set to 0.8, and FID is computed on the edited images.

\noindent\textbf{GRAF}~\cite{graf}.
We use the official implementation of \href{https://github.com/autonomousvision/graf}{GRAF}.
Since we observe that a slightly larger range of azimuth produces better results, we set the range of azimuth to $0^\circ$-$60^\circ$ and the elevation to $80^\circ$-$95^\circ$ in the training phase. However, when testing, we randomly sample the azimuth from the 30 degrees in the middle and fix the elevation to $90^\circ$. White background is set to false. Others are kept as the default ones.

\noindent\textbf{GRAF-RGBD}.
We modify the official implementation of GRAF by adding an extra discriminator to lead the learning of depth image generation. The new discriminator shares the same architecture as the original one, except that the input is changed to the one-channel depth image. We obtain the depth image by accumulating the depth with the help of sigma value. Other configurations are the same as those we used for GRAF.

\noindent\textbf{GIRAFFE}~\cite{GIRAFFE}.
We use the official implementation of \href{https://github.com/autonomousvision/giraffe}{GIRAFFE}.
The settings are kept the same as those for LSUN church. The depth image is obtained on the feature volume before 2D neural rendering by accumulating the depth with the help of sigma value. Since the output depth map is of resolution $16^2$, we then upsample it to the size of the input image through bilinear interpolation.

\noindent\textbf{$\pi$-GAN}~\cite{piGAN2021}.
We use the official implementation of \href{https://github.com/marcoamonteiro/pi-GAN}{$\pi$-GAN}.
The range of azimuth is set to $-30^\circ$-$+30^\circ$, and the elevation is fixed to $90^\circ$. During testing, the azimuth is sampled from the 30 degrees in the middle, and the elevation is set to $90^\circ$. White background is set to false. Other settings follow the configuration for CARLA dataset provided by the authors.

\noindent\textbf{$\pi$-GAN-RGBD}.
We make modifications on the original implementation of $\pi$-GAN. To incorporate the depth information into training, the input of the discriminator is changed from the three-channel RGB image into the four-channel RGBD image. Other hyperameters are kept the same as those we used for $\pi$-GAN.

\section{Rotation Consistency Loss}
With the images $\mathbf{I}^{f}_{rgbd,1}$ and $\mathbf{I}^{f}_{rgbd,2}$ generated under angles $\theta_1$ and $\theta_2$, we calculate the rotation consistency loss. $\mathbf{I}^{f}_{rgbd,2}$ will first be projected to the 3D space as a point cloud using the fixed camera intrinsic parameter $\mathbf{K}$. Then we rotate the point cloud around the central axis which passes through the center point of $xz$-plane and is parallel to $y$-axis. The rotation angle is the difference between $\theta_1$ and $\theta_2$. The rotation axis and the rotation angle form the rotation matrix $\mathbf{R}$, which is then used to transform the points accordingly. After the rotation, we get the new coordinates for each pixel in $\mathbf{I}^{f}_{rgbd,2}$. Then, we use the \textit{grid\_sample} function in PyTorch to query the RGB value in $\mathbf{I}^{f}_{rgbd,1}$ according to the new coordinates, which gives us the rotated image $\mathbf{I}^{f,rot}_{rgbd,1}$. We get a mask $\mathbf{M}$ through coordinate comparison simultaneously to filter out the out-of-boundary regions. Therefore, the output image $\mathbf{I}^{f,rot}_{rgbd,1}$ should be the same as $\mathbf{I}^{f}_{rgbd,2}$. The rotation consistency loss is then calculated between $\mathbf{I}^{f,rot}_{rgbd,1}\otimes \mathbf{M}$ and $\mathbf{I}^{f}_{rgbd,2}\otimes \mathbf{M}$, where $\otimes$ denotes element-wise multiplication.

\section{3D Visualization}

To visualize the point clouds of each synthesized scenes, we first project each generated RGBD images the 3D space as point clouds using a fixed camera intrinsic parameter $\mathbf{K}$. Then, we use ICP registration~\cite{rusinkiewicz2001efficient} implemented in Open3D~\cite{Zhou2018} for point cloud registration. Finally, we fuse these point clouds into one point cloud and show it from different viewpoints in Fig. 1 of the main paper.

\section{Depth Estimation}
\begin{figure}[!ht]
    \centering
    % \vspace{-2pt}
    \includegraphics[width=1.0\linewidth]{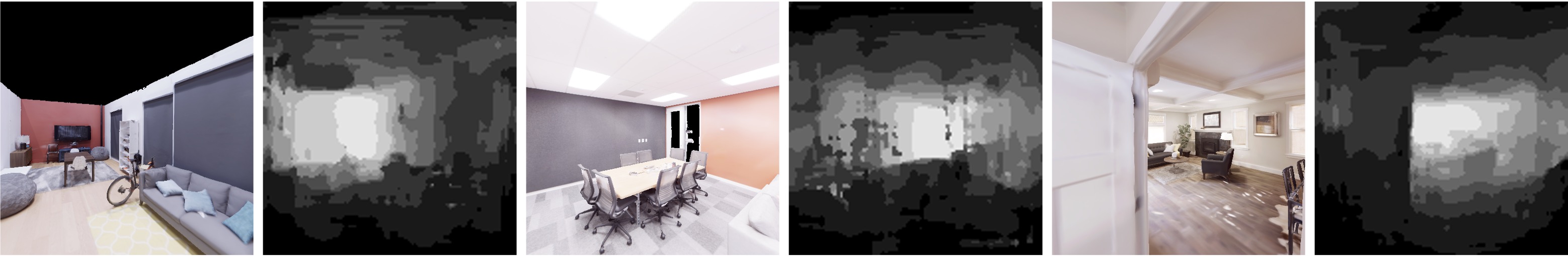}
    \vspace{-15pt}
    \caption{
        Estimated depth on the Replica dataset~\cite{replica}.
    }
    \label{fig:depth_estimation}
    \vspace{-13pt}
\end{figure}

Though depth estimation is not under the main scope of this work, we evaluate the switchable discriminator on Replica dataset~\cite{replica} to validate its transferability. We get 2.36 for 10-class cross-entropy error over 10K samples provided by \cite{GSN}. Some examples are visualized in \cref{fig:depth_estimation}.

\section{Additional Results}

\begin{figure*}[t]
    \centering
    \includegraphics[width=1.0\linewidth]{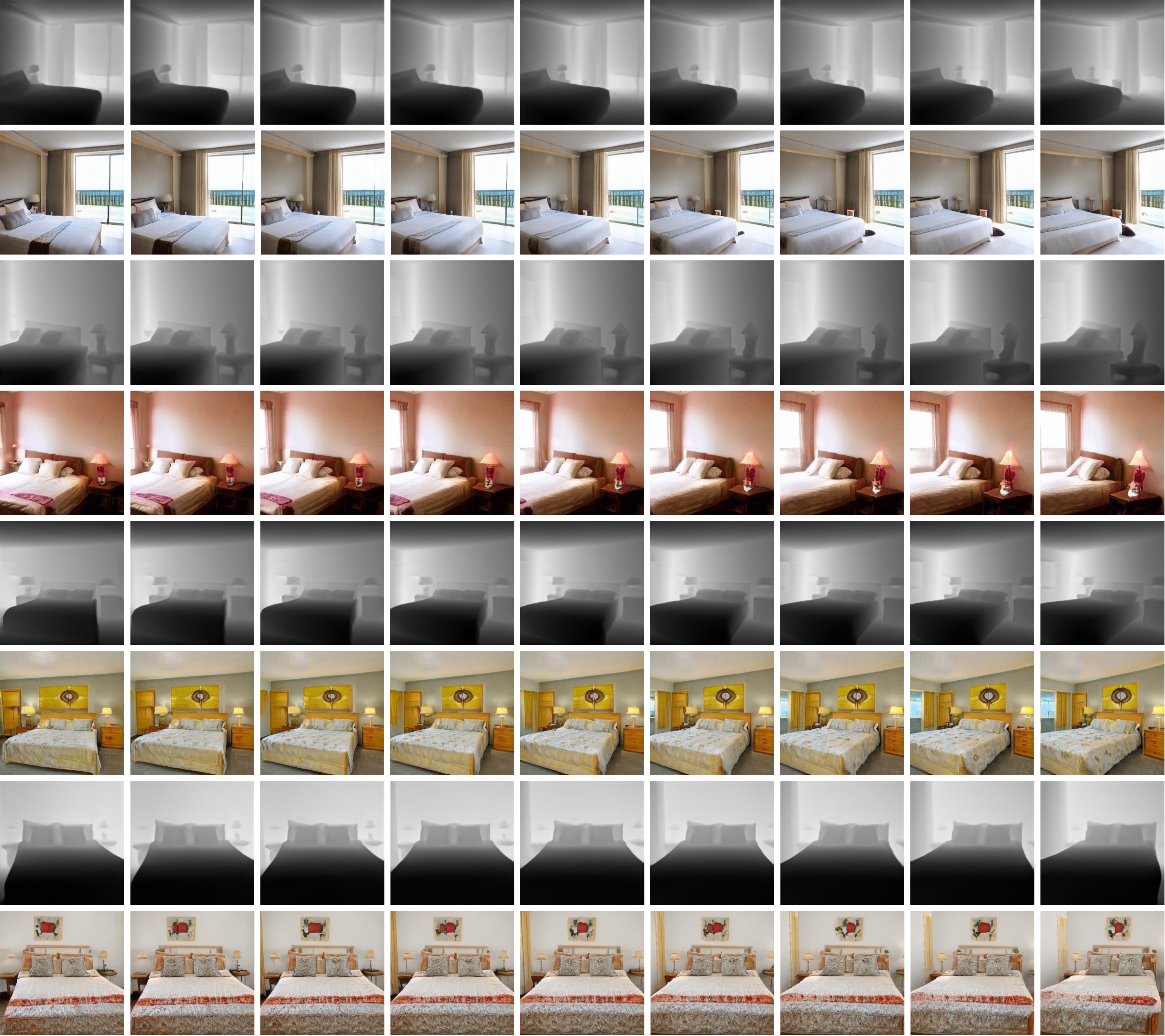}
    \caption{
        Results on LSUN bedrooms~\cite{yu2015lsun}.
    }
    \label{fig:supp_bed}
\end{figure*}

\begin{figure*}[t]
    \centering
    \includegraphics[width=1.0\linewidth]{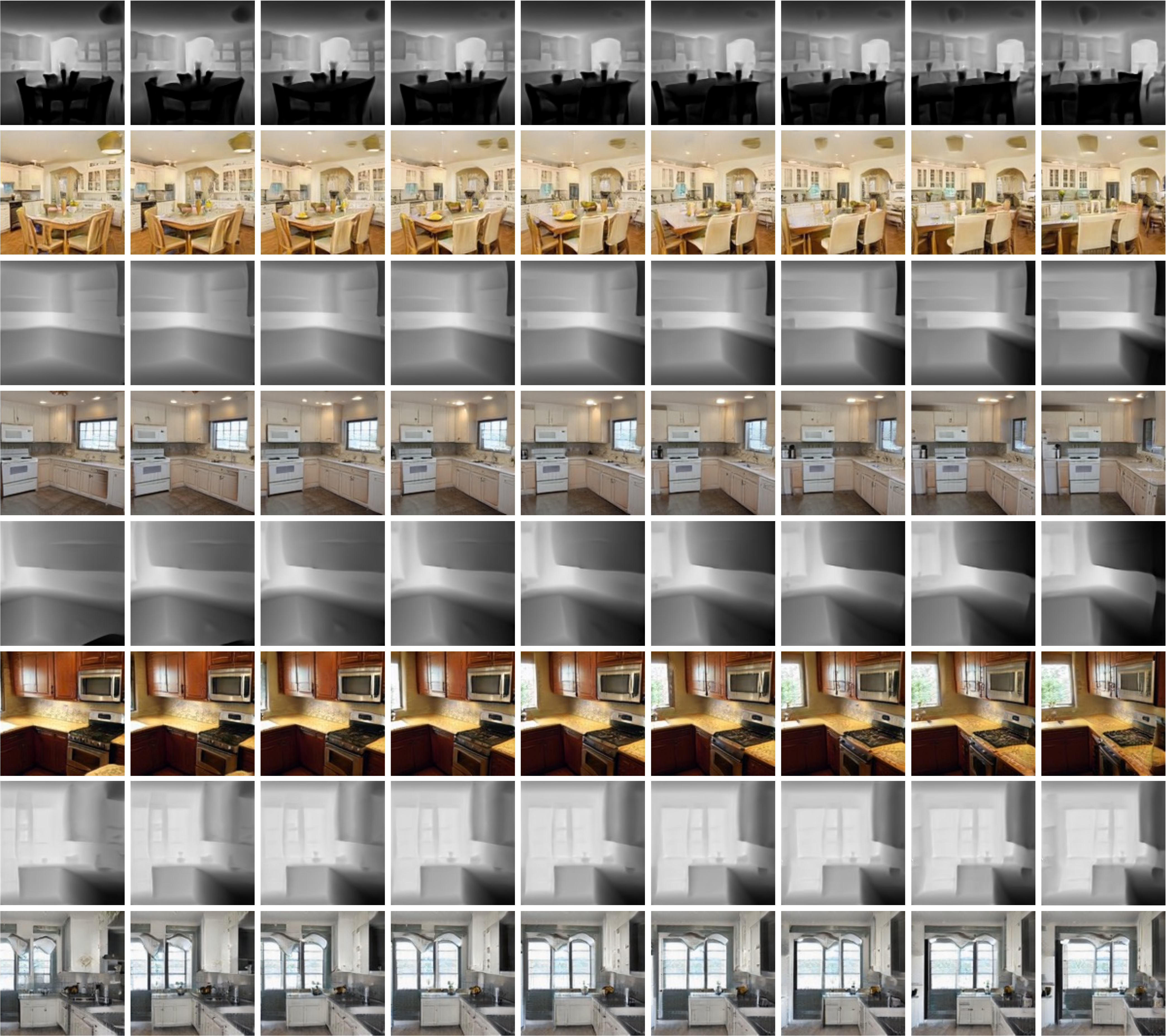}
    \caption{
        Results on LSUN kitchens~\cite{yu2015lsun}.
    }
    \label{fig:supp_kit}
\end{figure*}

More qualitative results are shown in \cref{fig:supp_bed} and \cref{fig:supp_kit}.
\href{https://youtu.be/RMmIso5Oxno}{Demo video} is also available to show the continuous 3D control achieved by our DepthGAN.

\end{document}